%% file: paper.tex
\documentclass[final]{cvpr}
\usepackage[utf8]{inputenc} 
\usepackage[T1]{fontenc}    
\usepackage[pagebackref=true,breaklinks=true,colorlinks,bookmarks=false]{hyperref}
\usepackage{url}            
\usepackage{booktabs}       
\usepackage{amsfonts}       
\usepackage{nicefrac}       
\usepackage{microtype}  
\usepackage{float} 

\usepackage{mathtools}
\usepackage{amssymb}
\usepackage{bbm}
\usepackage{multicol}
\usepackage{multirow}
\usepackage{graphicx}
\makeatletter
\@namedef{ver@everyshi.sty}{}
\makeatother
\usepackage{tikz}
\usepackage[export]{adjustbox}
\usepackage{caption}
\usepackage{thmtools}
\usepackage{thm-restate}
\usepackage{subfig}
\usepackage{etoc}
\usepackage{multibib}

\usepackage{enumitem,kantlipsum}
\usepackage[tight-spacing=true]{siunitx}
\usepackage{tablefootnote}
\usepackage{booktabs}
\usepackage{dsfont}
\usepackage{algcompatible}
\usepackage{gensymb}
\usepackage[hang,flushmargin]{footmisc}
\usepackage{lipsum}
\usepackage{color, colortbl}
\usepackage{cleveref}
\newcites{latex}{Appendix References}

\definecolor{Gray}{gray}{0.5}
\definecolor{gray}{rgb}{0.52, 0.52, 0.51}
\DeclareMathOperator*{\argmin}{arg\,min}
\DeclareMathOperator*{\argmax}{arg\,max}
\DeclarePairedDelimiterX{\infdivx}[2]{\bigg(}{\bigg)}{%
  #1\;\delimsize\|\;#2%
}
\usepackage{algorithm}
\newlength\myindent
\newcommand*{\referto}[1]{
 \refstepcounter{equation}%
 \tag*{\textcolor{blue}{[#1]} \hspace{0.1em} (\theequation)}%
}

\newcolumntype{P}[1]{>{\centering\arraybackslash}p{#1}}

\newcommand{\algorithmfootnote}[2][\footnotesize]{%
  \let\old@algocf@finish\@algocf@finish
  \def\@algocf@finish{\old@algocf@finish
    \leavevmode\rlap{\begin{minipage}{\linewidth}
    #1#2
    \end{minipage}}%
  }%
}
\newcommand\blfootnote[1]{%
  \begingroup
  \renewcommand\thefootnote{}\footnote{#1}%
  \addtocounter{footnote}{-1}%
  \endgroup
}
\def\cvprPaperID{6394}
\def\confYear{CVPR 2021} 
\begin{document}
\title{Greedy Hierarchical Variational Autoencoders for Large-Scale Video Prediction}
\author{
  Bohan Wu, Suraj Nair, Roberto Mart\'{i}n-Mart\'{i}n, Li Fei-Fei\footnote[2]{}, Chelsea Finn\footnote[2]{}\\
  Stanford University, Stanford, CA 94305 \\
  \tt\small \texttt{\{bohanwu, surajn, robertom, feifeili, cbfinn\}@cs.stanford.edu} \\
}
\maketitle
\begin{abstract}
A video prediction model that generalizes to diverse scenes would enable intelligent agents such as robots to perform a variety of tasks via planning with the model. However, while existing video prediction models have produced promising results on small datasets, they suffer from severe underfitting when trained on large and diverse datasets. To address this underfitting challenge, we first observe that the ability to train larger video prediction models is often bottlenecked by the memory constraints of GPUs or TPUs. In parallel, deep hierarchical latent variable models can produce higher quality predictions by capturing the multi-level stochasticity of future observations, but end-to-end optimization of such models is notably difficult. Our key insight is that greedy and modular optimization of hierarchical autoencoders can simultaneously address both the memory constraints and the optimization challenges of large-scale video prediction. We introduce Greedy Hierarchical Variational Autoencoders (GHVAEs), a method that learns high-fidelity video predictions by greedily training each level of a hierarchical autoencoder. In comparison to state-of-the-art models, GHVAEs provide 17-55\% gains in prediction performance on four video datasets, a 35-40\% higher success rate on real robot tasks, and can improve performance monotonically by simply adding more modules. Visualization and more details are at \url{https://sites.google.com/view/ghvae}.
\end{abstract}
\blfootnote{$\dagger$ Equal advising and ordered alphabetically.}
\input{1-intro}
\input{2-rw}
\input{3-ghvae}
\input{4-exps}
\input{5-conclusion}

\section*{Acknowledgements}
This work was supported in part by ONR grant N00014-20-1-2675. SN was supported by an NSF graduate research fellowship.

\bibliographystyle{IEEEtran}
\footnotesize\input{paper.bbl}

\newpage
\appendix
\onecolumn
\etocdepthtag.toc{mtappendix}
\etocsettagdepth{mtchapter}{none}
\etocsettagdepth{mtappendix}{subsubsection}
\tableofcontents
\normalsize
\section{Method}
\label{appendix:method}
\subsection{Memory Efficiency}
\subsubsection{GHVAE}
Because GHVAEs optimize each module with regard to image reconstruction, we must include in memory both the current module and some of the prior modules. Here, we briefly describe the memory savings of GHVAEs.
GHVAEs save GPU or TPU memory allocation by avoiding the need to store gradient information in previous modules during back-propagation. Specifically, for the encoder, intermediate activations and all gradients from the frozen modules no longer need to be stored in memory. For the decoder, the gradients of the activations will still need to be stored for backpropagation into the currently trained module. Table~\ref{memory} quantifies the amount of GPU or TPU memory saved for 1 to 6-module GHVAE models. This table indicates that the memory savings of a GHVAE model increases as the number of modules increases.
\begin{table}[h]
    \centering
    \begin{tabular}{|c|c|c|c|c|c|c|}
    \hline
        Model Parameter & \multicolumn{6}{c|}{Value} \\\hline
        Number of Modules $K$ & 1 & 2 & 3 & 4 & 5 & 6 \\\hline
        End-to-End Training Memory Usage (GB) & 3.44 & 4.63 & 5.79 & 13.57 & 19.99 & 28.34 \\
        Greedy Training Memory Usage (GB) & 3.44 & 3.46 & 4.23 & 9.20 & 13.60 & 17.05\\
        Memory Saved (\% Greedy Training Memory) & 0\% & 33.8\% & 36.9\% & 47.5\% & 47.0\% & 66.2\%\\
        \hline
    \end{tabular}
    \caption{\textbf{GPU or TPU Memory Usage of GHVAE Models.} All numbers are computed on a batch size of 1 per GPU, a rollout horizon of 10, two context frames, and $64 \times 64 \times 3$ image observations.}
    \label{memory}
\end{table}

\subsubsection{Other Methods}
While GHVAEs alleviate the challenge of training large-scale video prediction models in the face of GPU or TPU memory constraints, there are other ways of addressing this challenge, such as increasing the number of GPUs or TPUs (as opposed to increasing the memory capacity per GPU or TPU), having different examples on different GPUs, and allocating model weights across more than one GPUs. Our method is orthogonal and complementary to such directions. Also, while increasing the number of GPUs or TPUs can increase the training batch size, our method can still allow larger models to be trained even after batch size per GPU lowers to 1. 

It is also important to note that greedy training leads to higher optimization stability for GHVAEs in particular, as revealed in Ablation 2 of Table~\ref{tab:ablation2} in the main paper. Ablation 2 indicates that when GHVAEs are trained end-to-end from scratch, the model was unable to converge to any good performance in any single run compared to the greedy setting. GPU or TPU memory saving is only one of the benefits of performing greedy training.

\subsection{Intuition}
\label{appendix:intuition}
\begin{figure}[t]
    \centering
    \subfloat[][Current Image Observation]{\includegraphics[width=0.3\linewidth]{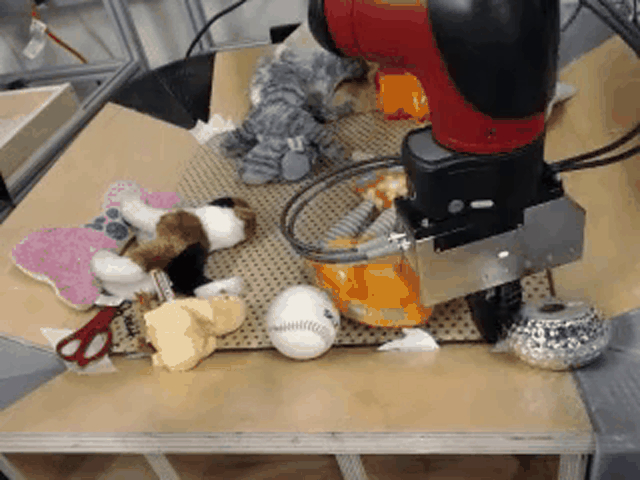}
    \label{robot-train}}\hspace{10pt}
    \subfloat[][Next Image Observation]{
    \includegraphics[width=0.3\linewidth]{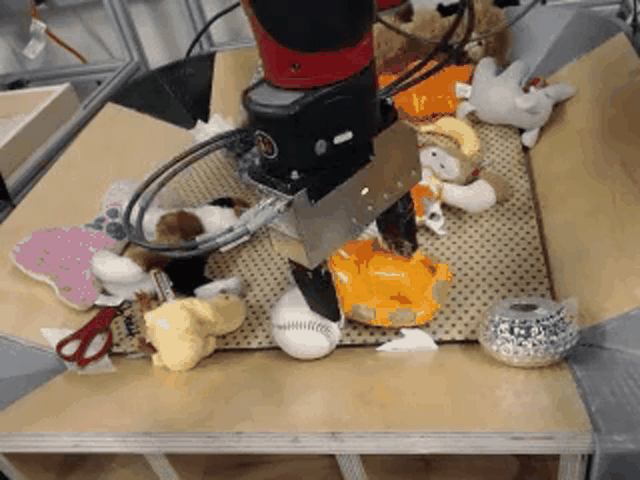}
    \label{robot-test}}
    \caption{\small \textbf{Example pair of current and next image in robotic manipulation.}}
    \label{intuition}
\end{figure}

In this section, we elaborate on the main paper's intuition on why it is important to capture the multi-level stochasticity of future observations in video prediction. Shown in Fig.~\ref{intuition} is an example of a current and next image observation from RoboNet. In action-conditioned video prediction for RoboNet, the video prediction model is given a four-dimensional vector $[dx, dy, dz, gripper]$, in which $dx, dy, dz$ denote the future end-effector translation from the current position, and $gripper$ is a binary integer for opening ($gripper=0$) or closing ($gripper=1$) the gripper. To accurately predict the next image observation, the video prediction model needs to precisely capture the end-effector position from the current monocular image, so that given the expected end-effector translation, the model can predict the new end-effector position and reconstruct all pixels that belong to the robot in the next image accordingly. The current end-effector position is considered a high-level visual feature that has inherent stochasticity because it is difficult to measure how long an inch is in this monocular image and therefore challenging to predict the precise pixel location of the robot in the next timestep. In addition, as the robot moves to a new position, the pixels currently occluded by the robot's arm will be revealed, and yet it is highly uncertain what is behind the robot's arm, let alone to predict these pixels for the next timestep. Concretely, there could be one or more objects behind the robot arm or zero objects. In the case where there are one or more objects, the ground truth texture and orientation of these objects are almost entirely occluded and unknown. These are the uncertainties around the low-level features in the image. In summary, multiple levels of uncertainty exist in the current image (from the high-level representation of end-effector position to the lower-level texture of the occluded objects and table background), therefore demanding the video prediction model to accurately model such multi-level stochasticity of future observations with hierarchical architectures.

As a side note, in the main paper, we posit that ``VAEs treat each dimension of a stochastic latent variable as independent''. Here, this statement refers to the case where the VAE uses a diagonal multivariate Gaussian distribution to model the latent variable distribution, which applies to GHVAEs as well.

\section{Experiments}
\label{appendix:experiments}
\subsection{Video Prediction}
In this section, we visualize qualitative results and discuss how we calculate each performance metric for video prediction, how we perform human evaluation using Amazon Mechanical Turk, and additional ablation studies.
\subsubsection{Visualizations}
\input{robonet}
\input{kitti}
\input{human}
\input{cityscapes}
\input{sweeping}
\input{wiping}
Figure~\ref{robonet},~\ref{kitti},~\ref{human},~\ref{cityscapes} exhibits example rollouts from video prediction methods reported in the main paper. Figure~\ref{sweeping} and~\ref{wiping} are the example rollouts from real-robot experiments: Pick\&Sweep and Pick\&Wipe tasks.

\subsubsection{Performance Evaluation}
Methodologies for calculating performance metrics are available at Table~\ref{tbl:metrics}. Note that these methodologies match those reported in prior works so that experiments conducted in this paper provide fair comparisons.
\begin{table}[h]
    \centering
    \resizebox{\textwidth}{!}{
    \begin{tabular}{|c|c|c|c|c|c|c|c|}
    \hline
        \multirow{2}{*}{Dataset} & Action-free & Batch & \# of Context & Rollout & GPU Memory & \# of  & FVD \\
        & / Action-conditioned & Size &Frames & Horizon & Usage (GB) & GPUs & Batch Size \\\hline
        RoboNet & Action-conditioned & 32 & 2 & 10 & 24 & 4 & 256\\
        KITTI & Action-free & 32 & 5 & 25 & 24 & 4 & 148\\
        Human3.6M & Action-free & 32 & 5 & 25 & 24& 4 & 256 \\
        CityScapes & Action-free & 128 & 2 & 28 & 16 & 8 & 256 \\
        Real-Robot Experiments & Action-conditioned & 140 & 2 & 10 & 24 & 4 & 256\\\hline
    \end{tabular}}
    \caption{GPU Memory Usage for All Experiments in Table~\ref{tab:svg} and Table~\ref{tab:vrnn}. All Convolutional Layers in the 6-Module GHVAE model for CityScapes are Downsized by 40\% to fit into 16GB GPU Memory for Fair Comparison.}
    \label{tbl:metrics}
\end{table}
\subsubsection{Human Evaluation}
For human evaluation, we provide 300 videos from both GHVAE and SVG' to Amazon Mechanical Turk workers in the form of 300 tasks. In each task, the workers are presented with three videos:  a video generated by GHVAE, a video generated by SVG', and the ground truth video. The worker does not know which video is generated by GHVAE or SVG', but do know which one is the ground truth video. In each task, the workers are asked to select the video that is more realistically similar to the ground truth video. These selections count as preferences. We then average all preferences and report results in Table~\ref{tab:svg} and~\ref{tab:vrnn}.

\subsubsection{Ablation for Encoder-Decoder Architectures}
In the early stages of our research, we have experimented with an alternative encoder-decoder architecture that expands or keeps the spatial dimension constant while reducing the channel dimension instead. The empirical performance of doing so significantly underperforms the current GHVAE architecture, which reduces spatial dimensions iteratively and compensates this dimensionality reduction by expanding the channel dimension. As mentioned in the paper, we hypothesize that reducing the spatial dimensions allows GHVAEs to perform better mean-field approximation in the deepest latent space.
\subsubsection{Ablation for Single vs. Multiple Latents}
In this section, we provide further intuition for the tradeoff between using single vs. multiple latent variables in a K-module GHVAE. Using multiple latent variables for GHVAE is an obvious option that we have empirically experimented with without satisfying results. Experimentally, when the GHVAE model uses all $K$ latent variables, the earlier latent variables provide suboptimal information and undesirably noisy signals to the overall network because of their inability to perform high-fidelity mean-field approximation when the spatial dimensions are large. This empirical phenomenon motivated us to only use the deepest latent variable in a GHVAE model. It is however important to note that using a single latent variable does not prevent GHVAEs from learning to accurately represent the multi-level stochasticity of future pixels. One can model such multi-level stochasticity using a single latent variable, provided that the decoders learn to appropriately project stochasticity from a succeeding layer to a preceding layer via non-linear transformation. In summary, we designed the GHVAE model to contain a single level of stochastic prediction, which is propagated through earlier deterministic layers to model multi-level stochasticity of future observations.

\subsection{Real Robot}
In this section, we elaborate on real-robot experimental setup and training data, visualizations of real-robot task execution and environments, and the random shooting planner we use to control the Franka robot.
\subsubsection{Setup}
In the first \textbf{Pick\&Wipe} task, the robot needs to pick a wiping tool (e.g. sponge, table cloth, etc.) up and wipe all objects off the plate for cleaning using the wiping tool. Each of the 20 trials contains different plates, objects, and wiping tools all unseen during training, and there could be at most two objects on the plate. The initial physical locations of the plate, the objects on the plate, and the robot itself are all randomized except that the robot is above the wiping tool. At the beginning of each trial, the wiping tool is \textit{not yet} in the robot's gripper, which makes the task more difficult. The task is considered successful if the robot picks the wiping tool up successfully and all objects are entirely wiped off the plate using the wiping tool within 50 timesteps. 

In the second \textbf{Pick\&Sweep} task, the robot is required to pick a sweeping tool (e.g. dustpan sweeper, table cloth, or dry sponge, etc.) up and sweep an object into the dustpan that is randomly placed in the bin. At the beginning of each trial, the sweeping tool is \textit{not yet} in the robot's gripper, which makes the task difficult. When a sweeping tool is not present in the scene, the robot then needs to sweep the object into the dustpan using its gripper. Each of the 20 trials contains different dustpans, objects, and sweeping tools all unseen during training. The physical location of the dustpan is uniformly random, and the object and the robot are also arbitrarily placed except that the robot is above the sweeping tool. The task is determined successful if the target object is swept into the dustpan within 50 timesteps. When a sweeping tool is indeed present, pushing the object into the dustpan using the robot's gripper will be considered a failure. Only pushing the object using the sweeping tool will be considered successful. This requires the video prediction methods to detect whether a tool was used for sweeping in the goal image and act accordingly in the physical task. 
\subsubsection{Training Data}
The video prediction models used for the real-robot experiments in this paper are not trained using the RoboNet dataset directly, but instead first pre-trained on RoboNet and then fine-tuned on a self-collected dataset of 5000 videos using the target Franka robot. Yet, this paper is about fitting video prediction models to large-scale datasets and this training scheme might seem to be contradicting with the main message. While the models can be trained directly on RoboNet, without fine-tuning on the 5000-video Franka dataset, the empirical task success rate is much lower for both GHVAE and SVG' on the target Franka environment due to unseen lighting conditions and camera viewpoint. On the other hand, if the models are only trained on the 5000-video dataset, the models easily overfit and fail to generalize to novel objects and tools. The purpose of large-scale video prediction is not to overfit a large dataset, but to learn powerful generalization such that the model can perform few-shot learning on the target environment using a small amount of data. Such a training scheme works in favor of learning large-scale video prediction, as opposed to defeating its purpose. Example environments for self-supervised training data collection are available at Fig.~\ref{fig:train-env}. 

The collection of training data is entirely self-supervised. Concretely, the robot randomly interacts with the training objects in the bin for 2-3 minutes in episodes of 20 timesteps, before pushing the objects from the corners to the center of the bin, so that object interaction remains frequent.
\subsubsection{Task Execution}
\input{sweeping-robot}
\input{wiping-robot}
Figure~\ref{sweeping-robot} and~\ref{wiping-robot} exhibit example Pick\&Sweep and Pick\&Wipe trials of real-robot task execution using the GHVAE and SVG' methods. Real-robot execution videos are at \url{https://sites.google.com/view/ghvae}.

\subsubsection{Task Diversity}
In Figure~\ref{fig:env}, we visualize more environments and tools used for real-robot tasks to reveal the diversity of the evaluation tasks. All objects used for evaluation are unseen during training.
\input{task-env}
\subsubsection{Planning}
\label{sec:randomshooting}
For simplicity, all real-robot experiments in this paper use a random shooting planner to optimize actions in visual foresight. Concretely, given a video prediction model and a goal image, we randomly sample a batch of 140 trajectories from the model and select the action sub-sequence for which the predicted images lead to the lowest L1 loss to the provided goal image. The robot replans after each execution of action sequences until the horizon of 50 timesteps is reached.

Concretely, the action space for the Franka robot has a dimension of 4 ($\mathcal{A} = \mathbb{R}^4$), which contains three scalars for the $[x, y, z]$ end-effector translation and one binary scalar for opening vs. closing its parallel-jaw gripper. Given the current image $x_{t}$, a goal image $g$, a sequence of $t$ context images $x_{1:t}$ and a sampled action sequence $a_{t:t+T-1}$, the sequence of frames predicted by the video prediction model $f$ is:
\begin{align}
    \hat{x}_{t'+1} = f(\hat{x}_{t'}, a_{t'}, x_{1:t})
\end{align}
where $t' \in [t, t+T-1]$, $\hat{x}_{t}=x_{t}$.

In practice, $T=10$ for the Franka robot, and we sample a batch of 140 action sequences $\{a_{t:t+T-1}^1, \ldots, a_{t:t+T-1}^{140}\}$ and predicted frames $\{\hat{x}_{t+1:t+T}^1, \ldots, \hat{x}_{t+1:t+T}^{140}\}$.

Next, we calculate the optimal length of action sequence $T^* \in [1, T]$, and the best action sequence index $b^* \in [1, 140]$ using the following equation:
\DeclarePairedDelimiter\abs{\lvert}{\rvert}
\begin{align}
    b^*, T^* = \argmin_{b \in [1, 140], T' \in [1, T]} \abs{\hat{x}^b_{t+T'} - g}
\end{align}

Finally, the best action sequence is then calculated as: $a_{1:T^*} = a^{b^*}_{1:T^*}$. The robot then executes this $T^*$-timestep action sequence and repeats this planning procedure.

\label{appendix:robot}
\section{Mathematical Proofs}
\label{appendix:proof}
\subsection{Proof of Theorem~\ref{theorem:lowerbound}}
\validity*
where $\mathcal{L}^{k}_{e2e}(x_{t+1})$ is GHVAE's ELBO for timestep $t+1$ when optimized end-to-end. More formally, $\mathcal{L}^{k}_{e2e}(x_{t+1})$ is $\mathcal{L}^{k}_{greedy}(x_{t+1})$ in Eq.~\ref{eq:elbo}, except that the VAE model $p^k \equiv p_{\mathcal{W}^{1 \ldots {k-1}, k}_{enc, dec, prior}}$ and the variational distribution $q^k \equiv q_{\mathcal{W}^{1 \ldots {k-1}, k}_{enc, post}}$.

\textbf{Proof.} Suppose $\mathcal{W}^{k^*}$ is the optimal parameters of the last module of a $k$-module GHVAE model:
\begin{align}
    \mathcal{W}^{k^*} = \argmax_{\mathcal{W}^k} \mathcal{L}_{greedy}^k(x_{t+1})
\end{align}
In other words:
\begin{align}
    \max_{\mathcal{W}^k} \mathcal{L}_{greedy}^k(x_{t+1}) = \mathcal{L}_{greedy}^k(x_{t+1}; \mathcal{W}^{k^*})
\end{align}
Therefore:
\begin{align}
    \log p(x_{t+1}) &\geq\max_{\mathcal{W}^{1 \ldots k}}  \mathcal{L}_{e2e}^k(x_{t+1}) \geq \mathcal{L}_{greedy}^k(x_{t+1}; \mathcal{W}^{k^*}) = \max_{\mathcal{W}^k} \mathcal{L}_{greedy}^k(x_{t+1})
\end{align}

\subsection{Proof of Theorem~\ref{theorem:monotonic}}
Recall that: 
\begin{align}
     \mathcal{L}^{k}_{greedy}(x_{t+1}) = \mathbb{E}_{q^k(z^k_{t+1} \mid x_{t+1})} \left[\log p^k(x_{t+1} \mid x_t, z_{t+1}^k)\right] - D_{KL}\infdivx{q^k(z_{t+1}^k \mid x_{t+1})}{p^k(z^k_{t+1} | x_t)}
     \label{eq:recall}
\end{align}
Steps in the following derivation that don't change from the previous step are in gray, while annotations are in blue.
\begin{align}
    & \log p(x_{t+1}) \geq \mathcal{L}^{k}_{greedy}(x_{t+1}) \referto{Variation Lower-Bound}\\
    &= \mathbb{E}_{q^k(z^k_{t+1} \mid x_{t+1})} \left[\log p^k(x_{t+1} \mid x_t, z^k_{t+1})\right] - D_{KL}\infdivx{q^k(z^k_{t+1} \mid x_{t+1})}{p^k(z^k_{t+1} \mid x_t)} \referto{Eq.~\ref{eq:recall}}\\
    &= \textcolor{Gray}{\mathbb{E}_{q^k(z^k_{t+1} \mid x_{t+1})}} \left[\log \int_{z^{k-1}_{t+1}} p^k(x_{t+1} \mid z^{k-1}_{t+1}, x_t, z^k_{t+1})p^k(z^{k-1}_{t+1} \mid x_t, z^k_{t+1})\frac{q^{k-1}(z^{k-1}_{t+1} \mid x_{t+1})}{q^{k-1}(z^{k-1}_{t+1} \mid x_{t+1})}\right] \nonumber\\
    &\quad \textcolor{Gray}{- D_{KL}\infdivx{q^k(z^k_{t+1} \mid x_{t+1})}{p^k(z^k_{t+1} \mid x_t)}}\referto{Algebra}\\
    &= \textcolor{Gray}{\mathbb{E}_{q^k(z^{k}_{t+1} \mid x_{t+1})}} \left[\log \mathbb{E}_{q^{k-1}(z^{k-1}_{t+1} \mid x_{t+1})} \left[\frac{p^k(x_{t+1} \mid z^{k-1}_{t+1}, x_t, z^{k}_{t+1})p^k(z^{k-1}_{t+1} \mid x_t, z^{k}_{t+1})}{q^{k-1}(z^{k-1}_{t+1} \mid x_{t+1})}\right]\right] \nonumber\\
    &\quad\textcolor{Gray}{- D_{KL}\infdivx{q^k(z^k_{t+1} \mid x_{t+1})}{p^k(z^k_{t+1} \mid x_t)}}\referto{Algebra}\\
    &\geq \textcolor{Gray}{\mathbb{E}_{q^k(z^{k}_{t+1} \mid x_{t+1})}} \Bigg[\mathbb{E}_{q^{k-1}(z^{k-1}_{t+1} \mid x_{t+1})} \Bigg[\log p^k(x_{t+1} \mid z^{k-1}_{t+1}, x_t, z^{k}_{t+1}) + \log \frac{p^k(z^{k}_{t+1} \mid x_t, z^{k-1}_{t+1})p^k(z^{k-1}_{t+1} \mid x_t)}{p^k(z^{k}_{t+1} \mid x_t)} \nonumber\\
    &\quad \textcolor{Gray}- \log q^{k-1}(z^{k-1}_{t+1} \mid x_{t+1})\Bigg]\Bigg] \textcolor{Gray}{-D_{KL}\infdivx{q^k(z^k_{t+1} \mid x_{t+1})}{p^k(z^k_{t+1} \mid x_t)}}\referto{Jensen's Inequality, Bayes' Rule}\\
    &= \textcolor{Gray}{\mathbb{E}_{q^k(z^{k}_{t+1} \mid x_{t+1})} \Bigg[\mathbb{E}_{q^{k-1}(z^{k-1}_{t+1} \mid x_{t+1})}} \Bigg[\log \frac{p^k(x_{t+1} \mid z^{k-1}_{t+1}, x_t)q^k(z^{k}_{t+1} \mid x_{t+1}, z^{k-1}_{t+1}, x_t)}{p^k(z^{k}_{t+1} \mid z^{k-1}_{t+1}, x_t)} \textcolor{Gray}{+ \log \frac{p^k(z^{k}_{t+1} \mid x_t, z^{k-1}_{t+1})p^k(z^{k-1}_{t+1} \mid x_t)}{p^k(z^{k}_{t+1} \mid x_t)}} \nonumber\\
    &\quad \textcolor{Gray}{- \log q^{k-1}(z^{k-1}_{t+1} \mid x_{t+1})} \Bigg]\textcolor{Gray}{\Bigg] - D_{KL}\infdivx{q^k(z^k_{t+1} \mid x_{t+1})}{p^k(z^k_{t+1} \mid x_t)}}\referto{Bayes' Rule}\\
    &= \textcolor{Gray}{\mathbb{E}_{q^k(z^{k}_{t+1} \mid x_{t+1})} \Bigg[\mathbb{E}_{q^{k-1}(z^{k-1}_{t+1} \mid x_{t+1})}} \Bigg[\log \frac{p^k(x_{t+1} \mid z^{k-1}_{t+1}, x_t)q^k(z^{k}_{t+1} \mid x_{t+1}, z^{k-1}_{t+1}, x_t)p^k(z^{k-1}_{t+1} \mid x_t)}{p^k(z^{k}_{t+1} \mid x_t)} \nonumber\\
    &\quad \textcolor{Gray}{- \log q^{k-1}(z^{k-1}_{t+1} \mid x_{t+1})}\Bigg]\Bigg] \textcolor{Gray}{-D_{KL}\infdivx{q^k(z^k_{t+1} \mid x_{t+1})}{p^k(z^k_{t+1} \mid x_t)\referto{Algebra}}}\\
    &= \textcolor{Gray}{\mathbb{E}_{q^k(z^{k}_{t+1} \mid x_{t+1})} \Bigg[\mathbb{E}_{q^{k-1}(z^{k-1}_{t+1} \mid x_{t+1})}} \Bigg[\log p^k(x_{t+1} \mid z^{k-1}_{t+1}, x_t) + \log p^k(z^{k-1}_{t+1} \mid x_t) - \log q^{k-1}(z^{k-1}_{t+1} \mid x_{t+1}) \nonumber\\
    &\quad+ \log q^k(z^{k}_{t+1} \mid x_{t+1}, z^{k-1}_{t+1}, x_t) - \log p^k(z^{k}_{t+1} \mid x_{t})\Bigg]\Bigg] \textcolor{Gray}{-D_{KL}\infdivx{q^k(z^k_{t+1} \mid x_{t+1})}{p^k(z^k_{t+1} \mid x_t)}}\referto{Algebra}\\
    &= \textcolor{Gray}{\mathbb{E}_{q^k(z^{k}_{t+1} \mid x_{t+1})}} \Bigg[\textcolor{Gray}{\mathbb{E}_{q^{k-1}(z^{k-1}_{t+1} \mid x_{t+1})}} \Bigg[\log p^{k-1}(x_{t+1} \mid z^{k-1}_{t+1}, x_t) + \log p^{k-1}(z^{k-1}_{t+1} \mid x_t) - \log q^{k-1}(z^{k-1}_{t+1} \mid x_{t+1}) \nonumber\\
    &\quad+ \log q^k(z^{k}_{t+1} \mid x_{t+1}, z^{k-1}_{t+1}, x_t)  \textcolor{Gray}{- \log p^k(z^{k}_{t+1} \mid x_{t})}\Bigg]\Bigg]  \textcolor{Gray}{-D_{KL}\infdivx{q^k(z^k_{t+1} \mid x_{t+1})}{p^k(z^k_{t+1} \mid x_t)}}\referto{$z^{k-1}_{t+1}$ is independent of $p^k$ given $p^{k-1}$}\\
    &= \textcolor{Gray}{\mathbb{E}_{q^k(z^{k}_{t+1} \mid x_{t+1})}} \left[\mathcal{L}^{k-1}(x_{t+1}) + \textcolor{Gray}{\mathbb{E}_{q^{k-1}(z^{k-1}_{t+1} \mid x_{t+1})}}\left[\log q^k(z^{k}_{t+1} \mid x_{t+1}, z^{k-1}_{t+1}, x_t) \textcolor{Gray}{- \log p^k(z^{k}_{t+1} \mid x_{t})}\right]\right] \nonumber\\
    &\quad \textcolor{Gray}{-D_{KL}\infdivx{q^k(z^k_{t+1} \mid x_{t+1})}{p^k(z^k_{t+1} \mid x_t)}}\referto{Eq.~\ref{eq:recall}}\\
    &= \mathcal{L}^{k-1}(x_{t+1}) + \textcolor{Gray}{\mathbb{E}_{q^k(z^{k}_{t+1} \mid x_{t+1})}} \left[\log q^k(z^{k}_{t+1} \mid x_{t+1}) \textcolor{Gray}{- \log p^k(z^{k}_{t+1} \mid x_{t})}\right] \textcolor{Gray}{ - D_{KL}\infdivx{q^k(z^k_{t+1} \mid x_{t+1})}{p^k(z^k_{t+1} \mid x_t)}}\referto{Remove conditionally independent variables, Algebra}\\
    &= \mathcal{L}^{k-1}(x_{t+1}) \referto{Algebra}
\end{align}
where $\mathcal{L}^{k-1} \in  \{\mathcal{L}^{k-1}_{greedy}, \mathcal{L}^{k-1}_{e2e}\}$ and $\mathcal{L}^{k}_{greedy}$ is initialized with the weights $\mathcal{W}^{1^* \ldots {k-1}^*}$. Notice that the proof above assumes action-free video prediction. The proof for action-conditioned video prediction is the same with every conditional variable $x_t$ in the proof above expanding into two joint conditional variables $x_t$ and $a_t$. For example, the term $p^k(x_{t+1} \mid x_t, z^k_{t+1})$ would be $p^k(x_{t+1} \mid x_t,  a_t, z^k_{t+1})$ instead.

\subsection{Clarification for Equation~\ref{eq:elbo}}
Note that while Eq.~\ref{eq:elbo} in the paper is an accurate mathematical form of GHVAE's ELBO, we have omitted $a_t$ in the term $\log p^k(x_{t+1} \mid x_t, z^k_{t+1})$ in this equation since GHVAE in practice only uses $a_t$ in the prior network. In other words, a more general form for Eq.~\ref{eq:elbo} is the following:
\begin{align}
     \mathcal{L}^{k}_{greedy}(x_{t+1}) = \mathbb{E}_{q^k(z^k_{t+1} \mid x_{t+1})} \left[\log p^k(x_{t+1} \mid x_t, a_t, z_{t+1}^k)\right] - D_{KL}\infdivx{q^k(z_{t+1}^k \mid x_{t+1})}{p^k(z^k_{t+1} | x_t, a_t)}
\end{align}

\section{Failure Case Analysis}
While a 6-Module GHVAE outperforms SVG' and Hier-VRNN, the model is still slightly underfitting RoboNet. We provide visualizations of failure examples in Figure~\ref{failure}. In this figure, the GHVAE model failed to accurately track the movement of the blue bowl. This indicates that the GHVAE model is still slightly underfitting on RoboNet. Given that such failure to track graspable object does not occur frequently for RoboNet, we hypothesize that this failure case is due to underfitting, and that training an 8-module, 10-module or 12-module GHVAE model can potentially tackle such failure case.

In addition, we hypothesize that a monocular image can cause partial observability to the video prediction problem. In Figure~\ref{failure} for example, without visually capturing the precise 3D locations of the robot and the blue bowl, it is difficult to tell whether the robot has successfully grasped the blue bowl and to predict the future motions of the blue bowl accordingly. Therefore, adding an $[x, y, z]$ state end-effector position vector or a second camera image from a different viewpoint (both are readily available information) to the GHVAE model can potentially resolve such a failure case.
\input{failure}

\end{document}

%% file: 1-intro.tex
\begin{figure}
    \centering
    \includegraphics[width=1.0\columnwidth]{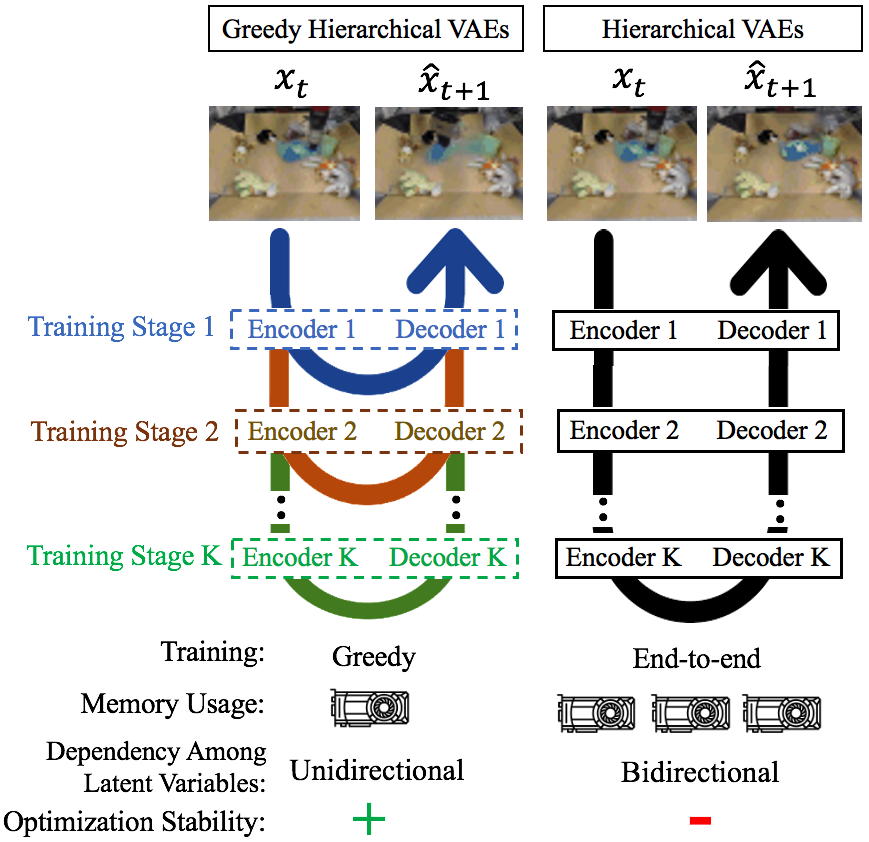}
    \caption{\small \textbf{Greedy Hierarchical Variational Autoencoders (GHVAEs).} Unlike traditional hierarchical variational autoencoders (VAEs), a GHVAE model trains each encoder-decoder module greedily using the frozen weights of the previously-trained modules. Greedy training circumvents fitting the entire model into memory and enables larger models to be trained within the same GPU or TPU memory. Further, greedy training improves the optimization stability of such a hierarchical model by breaking the bidirectional dependencies among individual latent variables. As a result, given the current image, $x_t$, GHVAE predicts a more accurate next image, $\hat{x}_{t+1}$, than a hierarchical VAE. Each module is optimized sequentially, and all modules are used at test time. }
    \label{fig:intro}
    \vspace{-15pt}
\end{figure}

\section{Introduction}
\label{s1_intro}

A core aspect of intelligence is the ability to predict the future. Indeed, if equipped with an accurate video prediction model, an intelligent agent such as a robot may be able to perform a variety of tasks using raw pixel inputs. For example, algorithms such as visual foresight~\cite{ebert2018visual} can leverage an action-conditioned video prediction model to plan a sequence of actions that accomplish the desired task objective. Importantly, such video prediction models can in principle be trained with broad, unlabeled datasets, and building methods that can learn from large, diverse offline data is a recipe that has seen substantial success in visual~\cite{deng2009imagenet} and language~\cite{brown2020language} understanding.
However, learning an accurate video prediction model from large and diverse data remains a significant challenge. The future visual observations of the world are hierarchical~\cite{palmer1977hierarchical}, high-dimensional, and uncertain, demanding the model to accurately represent the multi-level stochasticity of future pixels, which can include both low-level features (e.g. the texture of a table as it becomes unoccluded by an object) and higher-level attributes (e.g. how an object will move when touched), such as the top images in Fig.~\ref{fig:intro}.

To capture the stochasticity of the future, prior works have proposed a variety of stochastic latent variable models~\cite{babaeizadeh2017stochastic,lee2018stochastic,svg}.
While these methods generated reasonable predictions for relatively small video prediction datasets such as the BAIR robot pushing dataset~\cite{ebert2018robustness}, they suffer from severe underfitting in larger datasets in the face of practical GPU or TPU memory constraints~\cite{villegas2019high}. On the other hand, while hierarchical variational autoencoders (VAEs) can in principle produce higher-quality predictions by capturing multiple levels of stochasticity, the bidirectional dependency between individual hierarchical latent variables (higher-level variables influence the lower level and vice versa) potentially creates an unsolved problem of optimization instability as the number of hierarchical latent variables in the network increases~\cite{sonderby2016train,Castrejon_2019_ICCV}. 

The key insight of this work is that greedy and modular optimization of hierarchical autoencoders can simultaneously address both the memory constraints and the optimization challenges of learning accurate large-scale video prediction. On one hand, by circumventing end-to-end training, greedy machine learning allows sequential training of sub-modules of the entire video prediction model, enabling much larger models to be learned within the same amount of GPU or TPU memory. On the other hand, optimizing hierarchical VAEs in a greedy and modular fashion breaks the bidirectional dependency among individual latent variables. As a result, these variables can remain stable throughout the entire training process, resolving the typical instability of training deep hierarchical VAEs.

With this key insight, this paper introduces Greedy Hierarchical VAEs (``GHVAEs'' hereafter) (Fig.~\ref{fig:intro})-- a set of local latent VAE modules that can be sequentially stacked and trained in a greedy, module-by-module fashion, leading to a deep hierarchical variational video prediction model that in practice admits a stable optimization and in principle can scale to large video datasets.
As evaluated in Section~\ref{sec:exp}, GHVAEs outperform state-of-the-art video prediction models by 17-55\% in FVD score~\cite{unterthiner2018towards} on four different datasets, and by 35-40\% success rate on two real robotic manipulation tasks when used for planning.
In addition, our empirical and theoretical analyses find that GHVAE's performance can improve monotonically as the number of GHVAE modules in the network increases. In summary, the core contribution of this work is the use of greedy machine learning to improve both the optimization stability and the memory efficiency of hierarchical VAEs, leading to significant gains in both large-scale video prediction accuracy and real robotic task success rates.

%% file: 2-rw.tex
\begin{figure*}
    \centering
    \includegraphics[width=\linewidth]{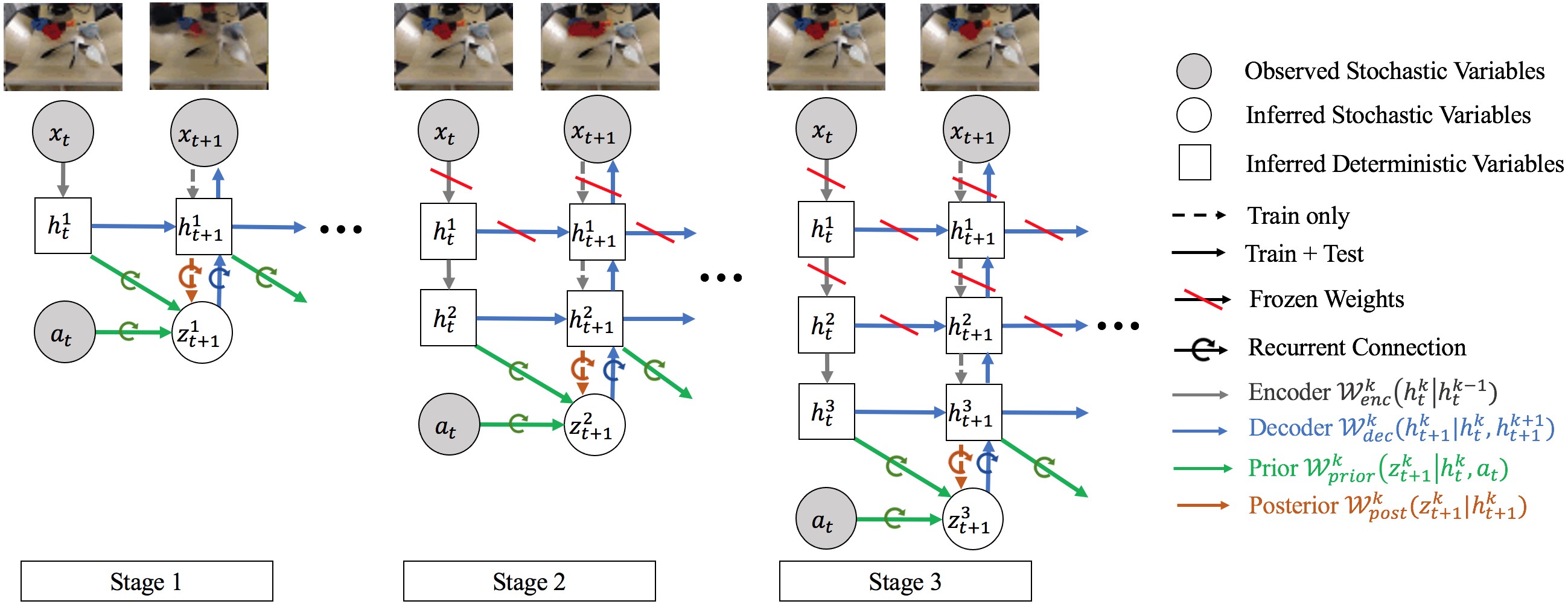}
    \caption{\small \textbf{Training procedure and architecture for a three-module GHVAE.} In Stage 1, all first-module weights ($\mathcal{W}^1_{enc}, \mathcal{W}^1_{dec}, \mathcal{W}^1_{prior}, \mathcal{W}^1_{post}$)
    are trained end-to-end. In Stage 2, all weights from the first module are frozen and the second module is trained. In Stage 3, all first and second-module weights are frozen, and only the third module is trained, so on and so forth. The video prediction quality for $x_{t+1}$ improves as more modules are added. The legends in the figure denote the four components in each GHVAE module (encoder, decoder, prior, and posterior) and whether each component is frozen (tilted red bars) or used only for training and not at test time (dashed as opposed to solid lines). To limit the number of spatial dimensions that requires prediction from the prior network, only the prior and posterior in the final, $K^{th}$ GHVAE module are used. The action $a_t$ is included in action-conditioned video prediction and excluded in action-free video prediction. }
    \label{fig:train}
    \vspace{-5mm}
\end{figure*}
\section{Related Work}
\noindent \textbf{The Underfitting Challenge of Large-Scale Video Prediction.} Resolving the underfitting challenge of large-scale video prediction can lead to powerful generalization in visual foresight~\cite{boots2014learning,finn2017deep,kalchbrenner2017video,ebert2017self,ebert2018robustness,ebert2018visual,xie2019improvisation,paxton2019visual,nair2019hierarchical,nair2020trass}, which performs model-based robotic control~\cite{polydoros2017survey,massoud2009model,zhang2019solar} via action-conditioned video prediction~\cite{oh2015action,hirose2019vunet,hirose2019deep,nunes2020action}.
Initially, video prediction~\cite{mathieu2016deep,lotter2016deep,liang2017dual,xu2018video,hsieh2018learning,reda2018sdc,xu2018structure,Castrejon_2019_ICCV,ye2019compositional} has been tackled by a deterministic model~\cite{walker2015dense,finn2016unsupervised,jia2016dynamic,xue2016visual,walker2016uncertain,liang2017dual,byravan2017se3,vondrick2017generating,van2017transformation,liu2017video,chen2017video,lu2017flexible}.
VAEs were later adopted to model the stochasticity of future visual observations~\cite{shu2016stochastic,babaeizadeh2017stochastic,wichers2018hierarchical,franceschi2020stochastic}. Nevertheless, modeling the stochasticity of the real world using a trajectory-based latent variable model leads to blurry predictions inadvertently. This problem is then addressed by two lines of orthogonal work-- VAE-GANs~\cite{lee2018stochastic} and timestep-based latent variable models~\cite{svg}. While these methods resolve blurry predictions in small-scale video datasets such as the BAIR robot pushing dataset~\cite{ebert2018robustness}, they suffer from severe underfitting in large-scale, multi-domain, or multi-robot datasets, such as RoboNet~\cite{dasari2019robonet} and RoboTurk~\cite{mandlekar2018roboturk}.
In parallel, Villegas et al.~\cite{villegas2019high} validate that higher model capacity leads to greater prediction fidelity.
This raises the question of how to learn larger models to meet the underfitting challenge of large-scale video prediction.
On the other hand, Castrejon et al.~\cite{Castrejon_2019_ICCV} apply dense connections to hierarchical VAEs to address the optimization challenge of fitting hierarchical variational video prediction models. While this work outperforms the state-of-the-art in relatively small video datasets, it was unable to scale its hierarchical VAE up substantially due to deep optimization problems~\cite{sonderby2016train,Castrejon_2019_ICCV}. 
Other works have also attempted to address the underfitting challenge of large-scale video prediction through other angles. For example, one line of work attempts to represent pixels as discrete as opposed to continuous distributions~\cite{van2016conditional,Salimans2017PixeCNN}.
Other works predict forward alternative quantities such as object-centric representations~\cite{janner2018reasoning,greff2019multi,watters2019cobra,veerapaneni2020entity,kipf2019contrastive,engelcke2019genesis} and goal-centric representations~\cite{nair2020goal}. Unlike these approaches, our method scales to large real-world video datasets without requiring additional inductive biases.

\noindent \textbf{Greedy Machine Learning}. Greedy machine learning~\cite{verbeek2003efficient,hinton2006fast,bengio2007greedy,haarnoja2018latent,belilovsky2019greedy,Malinowski_2020_CVPR} was first introduced to provide a good weight initialization for deep networks to escape bad local optima during end-to-end back-propagation.
As originally proposed, each greedy module of a deep network is stacked on top of the preceding greedy module and trained locally based on the features extracted from the preceding module. 
Subsequently, greedy machine learning has been applied to pre-training good feature extractors and stacked autoencoders~\cite{vincent2010stacked,masci2011stacked,zhang2014coarse,kumar2014static,ijjina2016classification,singh2018deep} for downstream tasks in vision, sound, and language~\cite{qi2014robust,lowe2019putting,lowe2019greedy}. Trained via self-supervised learning, these feature extractors and autoencoders excelled at capturing and preserving time-invariant information in sequential data such as videos.
In contrast, we propose a video prediction method that uses a hierarchy of latent variables to explicitly model \textit{time-variant} information about the future. Finally, greedy training of generative adversarial networks (GANs) is proposed to generate high-quality, high-resolution single-images~\cite{karras2017progressive}. Unlike these prior works, we propose a greedy approach to training large-scale video prediction models that simultaneously addresses the memory constraints and the optimization challenges of hierarchical VAEs.

\noindent \textbf{Hierarchical Variational Autoencoders}. Hierarchical~\cite{sonderby2016ladder} and sequential VAEs~\cite{zhao2017towards} were recently introduced to improve generative modeling in various vision tasks such as video prediction~\cite{Castrejon_2019_ICCV} and image generation~\cite{vahdat2020nvae}. They are known to have optimization challenges~\cite{sonderby2016train}, mainly due to the bidirectional dependency among the individual latent variables. When optimized end-to-end, the hierarchical VAE needs to keep each latent variable useful for the video prediction task at hand throughout the entire training process, while preserving the dependent relationships among these variables simultaneously. To this end, previous works introduced a variety of inductive biases such as dense connections~\cite{Castrejon_2019_ICCV}, ladder structures~\cite{zhao2017learning}, bidirectional inference~\cite{maaloe2019biva}, progressive lossy compression~\cite{ho2020denoising,song2020denoising}, and spectral regularization~\cite{vahdat2020nvae} to alleviate such optimization difficulties specific to hierarchical VAEs.
These approaches have largely been successful in the context of image generation, while we study the more difficult video prediction problem. Unlike these approaches, we propose a greedy training scheme that significantly alleviates the optimization challenges of conditional hierarchical VAEs.

%% file: 3-ghvae.tex
\section{Greedy Hierarchical VAEs (GHVAEs)}
\noindent \textbf{Overview.} To develop an expressive yet stably optimized video prediction model, we introduce Greedy Hierarchical VAEs (Fig.~\ref{fig:train}), which are locally optimized VAE modules that can be stacked together sequentially to incrementally add capacity to the model. To train a stack of modules without needing to fit the entire model into memory, each module is optimized locally using the frozen weights of the previously-trained modules.
Concretely, a GHVAE model has multiple GHVAE modules. Each GHVAE module has four convolutional sub-networks: an encoder, a decoder, a prior network, and a posterior inference network. In the remainder of this section, we overview mathematical notation, describe each of these model components in detail, derive the training objective for each module as a variational lower bound, and theoretically analyze the implications of greedy training. \\

\noindent \textbf{Notation.} This paper uses $K$ to denote the total number of GHVAE modules in the network, $\mathcal{W}^k, k \in [1, K]$ to denote the $k^{th}$ GHVAE module, $\mathcal{W}^k = \{\mathcal{W}^k_{enc}, \mathcal{W}^k_{dec}, \mathcal{W}^k_{prior}, \mathcal{W}^k_{post}\}$ to denote the $k^{th}$ module's encoder, decoder, prior network, and posterior inference network respectively, $x_t \in \mathcal{X} = \mathbb{R}^{H^0 \times W^0 \times C^0}$ to represent the 
RGB image observation (height $H^0$, width $W^0$, channel $C^0=3$) at the current timestep $t$, $h^k_t \in \mathcal{H}^k = \mathbb{R}^{H^k \times W^k \times C_\mathcal{H}^k}$ to denote the hidden variable encoded by the $k^{th}$ module for the current timestep $t$, $z^k_{t+1} \in \mathcal{Z}^k = \mathbb{R}^{H^k \times W^k \times C_\mathcal{Z}^k}$ to denote the $k^{th}$ stochastic latent variable used to explicitly model the stochasticity of the future observation at timestep $t+1$, $a_t \in \mathcal{A}$ to denote the agent's action at the current timestep $t$ in the case of action-conditioned video prediction, and $T$ to denote the model's roll-out horizon during training. \\

\noindent \textbf{Encoder.} Shown as grey downward arrows in Fig.~\ref{fig:train}, the $K$ encoders in a GHVAE model incrementally map from $x_t$ to $h^K_t$ and serve as part of both the VAE model and the posterior inference network. For the encoder design, it is important to recall that VAEs treat each dimension of a stochastic latent variable as independent (i.e. the mean-field approximation). However, convolutional embeddings of images contain significant spatial correlations due to the low frequency of natural images, violating this approximation. To mitigate this challenge, we design the encoder architecture to incrementally compress the spatial dimensions of the embeddings while simultaneously significantly expanding the channel dimensions of the embeddings. This allows the model, at its deepest layer, to store plenty of information (including spatial information) without strongly-correlated dimensions. Concretely, the $k^{th}$ encoder $\mathcal{W}_{enc}^k$ maps from $h_t^{k-1}$ to $h_t^k$ (except for the first encoder $\mathcal{W}_{enc}^0$, which maps $x_t$ to $h_t^1$), and incrementally compresses the height and width, $H^{k}<H^{k-1}$, $W^{k}<W^{k-1}$, while expanding the channels $C_\mathcal{H}^k>C_\mathcal{H}^{k-1}$.\\

\noindent \textbf{Decoder.} Shown as blue arrows in Fig.~\ref{fig:train}, the $K$ decoders in a GHVAE model incrementally map from the deepest stochastic latent variable $z^K_{t+1}$ back to $x_{t+1}$ to predict the next image. Since encoding significant information in stochastic latent variables is difficult, we aim to allow the stochastic latent variables to only capture new information about the future that is absent from the past. In other words, any partial information of the future that exists in $h_t^k$ does not need to be predicted and thus should not be contained in $z^k_{t+1}$. Hence, the decoder in the deepest latent space, $\mathcal{W}_{dec}^K$, takes as input both $h_t^K$ and the posterior latent variable $z_{t+1}^K$, so that the network can borrow information directly from the past. Similarly, each decoder $\mathcal{W}_{dec}^k \in \{\mathcal{W}_{dec}^1 \ldots \mathcal{W}_{dec}^{K-1}\}$ takes as input both $h_t^{k}$ and $h^{k+1}_{t+1}$ and predicts $h^{k}_{t+1}$ (except for $\mathcal{W}_{dec}^1$, which predicts $x_{t+1}$). Mirroring the encoders, these decoders incrementally expand the height and width, while compressing the channels. \\

\noindent \textbf{Prior Network.} Shown as green arrows in Fig.~\ref{fig:train}, the prior network $\mathcal{W}_{prior}^k$ maps $h^k_t$ and $a_t$ to the mean and variance of a diagonal Gaussian distribution for $z^k_{t+1}$ to model the stochasticity of future observations. The prior network is recurrent-convolutional and used both at train and test time. Empirically, using all $K$ stochastic latent variables $z^1_{t+1} \ldots z^K_{t+1}$ leads to excessive stochasticity and degrades performance as the number of GHVAE modules increases. Therefore, one key design choice is that while a $K$-module GHVAE uses all $K$ stochastic latent variables \textit{during training} (i.e., $z^{1\ldots K}_{t+1}$, one for each module) to sequentially learn the multi-level stochasticity of future observations, only the latent variable at the deepest level, $z^K_{t+1}$, is used \textit{at test time} and requires prediction from the prior network. This greedy training strategy allows each decoder to propagate uncertainty from the deepest layer to the shallower layers, and ultimately back to the pixel space. As a result, GHVAEs can implicitly model the multi-level stochasticity of future observations without explicitly using multiple stochastic latent variables \textit{at test time}, and can maximally compress the latent space spatially module-by-module such that $h^K_t$ and $z^K_{t+1}$ contain as few spatial dimensions as possible. Because the deepest encoder will have the fewest spatial dimensions, the only stochastic latent variable $z^K_{t+1}$ will have the least spatial correlations.\\

\noindent \textbf{Posterior Inference Network}. Although the encoder and decoder have minimized spatial dimensions in the deepest hidden layer $h^K$, the encoding process has produced a high channel dimension $C^K_\mathcal{H}$ for $h^K$. To improve the quality of prediction by the prior network, the channels in $h^{K}$ may need to be downsized to reduce the required output dimensions of the prior network. Hence, shown as brown arrows in Fig.~\ref{fig:train}, the posterior inference network maps the current module's hidden variable $h^k_{t+1}$ to the mean and variance of a diagonal Gaussian distribution over the stochastic latent variable $z^k_{t+1}$. When modules are added, a new posterior inference network and a new prior network for the new latent space are trained based on the latest module's representation. $z_{t+1}^k$ is a posterior latent variable, since both $h^k_{t+1}$ and $z^k_{t+1}$ are encoded from the ground truth future observation $x_{t+1}$ as opposed to the predicted next observation. For this reason, the recurrent-convolutional posterior network is only available at train time and not used for inference at test time. \\

\begin{table*}[t]
\centering
\caption{\small GHVAE vs. SVG' video prediction test performance (mean $\pm$ standard error). GHVAE outperforms SVG' on all datasets across all metrics. ``Human'' denotes human preferences between the two methods.} 
\label{tab:svg}
\small
\begin{tabular}{|c|c|c|c|c|c|c|}
\hline
\multirow{2}{*}{Dataset} & \multirow{2}{*}{Method} &   \multicolumn{5}{c|}{Video Prediction Test Performance} \\ 
& & FVD $\downarrow$ & 
PSNR $\uparrow$ & SSIM $\uparrow$ & LPIPS $\downarrow$ & Human\\ \hline
\multirow{2}{*}{RoboNet} & GHVAEs & \textbf{95.2$\pm$2.6}  & \textbf{24.7$\pm$0.2} &  \textbf{89.1$\pm$0.4} & \textbf{0.036$\pm$0.001} & \textbf{92.0\%}\\
 & SVG'~\cite{villegas2019high} &  123.2$\pm$2.6 & 23.9$\pm$0.1 & 87.8$\pm$0.3 & 0.060$\pm$0.008 & 8.0\%\\\hline
\multirow{2}{*}{KITTI} & GHVAEs & \textbf{552.9$\pm$21.2} & \textbf{15.8$\pm$0.1} & \textbf{51.2$\pm$2.4} & \textbf{0.286$\pm$0.015} & \textbf{93.3\%}\\
 & SVG'~\cite{villegas2019high} & 1217.3~\cite{villegas2019high}  & 15.0~\cite{villegas2019high} & 41.9~\cite{villegas2019high}  & 0.327$\pm$0.003 & 6.7\%\\\hline
\multirow{2}{*}{Human3.6M} &  GHVAEs & \textbf{355.2$\pm$2.9} & \textbf{26.7$\pm$0.2} & \textbf{94.6$\pm$0.5}  & \textbf{0.018$\pm$0.002} & \textbf{86.6\%}\\
 & SVG'~\cite{villegas2019high} & 429.9~\cite{villegas2019high} & 23.8~\cite{villegas2019high} &88.9~\cite{villegas2019high} & 0.028$\pm$0.006 & 13.4\% \\\hline
\end{tabular}
\vspace{-10px}
\end{table*}

\noindent \textbf{Optimization.} In this section, we use $p^k$ to denote the VAE model and $q^k$ to denote the variational distribution. The encoder, the decoder, and the prior network are all part of the model $p^k$, and both the encoder and the posterior inference network are part of $q^k$. The training process of a $K$-module GHVAE model is split into $K$ training phases, and only the $k^{th}$ GHVAE module is trained during phase $k$, where $k \in [1, K]$. GHVAE's training objective for the $k^{th}$ module is: 
\begin{align}
\max_{\mathcal{W}^k} \sum_{t=0}^{T-1} \mathcal{L}^{k}_{greedy}(x_{t+1})
\label{eq:greedy}
\end{align}
where $\mathcal{L}^{k}_{greedy}(x_{t+1})$ is GHVAE's Evidence Lower-Bound (ELBO) with respect to the current module $\mathcal{W}^k$ at timestep $t+1$:
\begin{align}
    \mathcal{L}^{k}_{greedy}&(x_{t+1})=\mathbb{E}_{q^k(z^k_{t+1} \mid x_{t+1})} [\log p^k(x_{t+1} \mid x_t, z_{t+1}^k)] \nonumber\\
    &- D_{KL}\infdivx{q^k(z_{t+1}^k \mid x_{t+1})}{p^k(z^k_{t+1} | x_t, a_t)}
    \label{eq:elbo}
\end{align}
where $p^k \equiv p_{\mathcal{W}^{1^* \ldots {k-1}^*, k}_{enc, dec, prior}}$, $q^k \equiv q_{\mathcal{W}^{1^* \ldots {k-1}^*, k}_{enc, post}}$, and $\mathcal{W}^{1^* \ldots {k-1}^*}$ are the frozen, greedily trained weights of all preceding GHVAE modules.

To improve training stability, we use a fixed standard deviation for the posterior latent variable distribution $q^k(z_{t+1}^k \mid x_{t+1})$ in the KL divergence term in Eq.~\ref{eq:elbo}. \\

\noindent \textbf{Theoretical Guarantees}. GHVAE's ELBO manifests two theoretical guarantees. \textbf{1) ELBO Validity:} Sequentially optimizing each GHVAE module in the network is equivalent to maximizing a \textit{lower-bound} of the ELBO for training all GHVAE modules end-to-end. This suggests that GHVAE's ELBO is \textit{valid}: 

\begin{restatable}[ELBO Validity]{thm}{validity}
For any $k \in \mathbb{Z}^+$ and any set of frozen, greedily or end-to-end trained weights $\mathcal{W}^{1^* \ldots {k-1}^*}$,
\begin{align}
    \log p(x_{t+1}) &\geq\max_{\mathcal{W}^{1 \dots k-1, k}} \mathcal{L}^{k}_{e2e}(x_{t+1}) \nonumber\\
    &\geq \max_{\mathcal{W}^k} \mathcal{L}^{k}_{greedy}(x_{t+1})
\end{align}
\label{theorem:lowerbound}
\end{restatable}
where $\mathcal{L}^{k}_{e2e}(x_{t+1})$ is GHVAE's ELBO for timestep $t+1$ when optimized end-to-end. More formally, $\mathcal{L}^{k}_{e2e}(x_{t+1})$ is $\mathcal{L}^{k}_{greedy}(x_{t+1})$ in Eq.~\ref{eq:elbo}, except that the VAE model $p^k \equiv p_{\mathcal{W}^{1 \ldots {k-1}, k}_{enc, dec, prior}}$ and the variational distribution $q^k \equiv q_{\mathcal{W}^{1 \ldots {k-1}, k}_{enc, post}}$.\\

\noindent \textbf{2) Monotonic Improvement:} Adding more modules can only raise (as opposed to lower) GHVAE's ELBO, which justifies and motivates maximizing the number of modules in a GHVAE model: 
\begin{restatable}[Monotonic Improvement]{thm}{monotonic}
For any $k \in \mathbb{Z}^+$ and any set of frozen, greedily or end-to-end trained weights $\mathcal{W}^{1^* \ldots {k-1}^*}$,
\begin{align}
     \log p(x_{t+1}) &\geq\mathcal{L}^{k}_{greedy}(x_{t+1}; \mathcal{W}^{1^* \ldots {k-1}^*}) \nonumber\\
     &\geq \mathcal{L}^{k-1}(x_{t+1}; \mathcal{W}^{1^* \ldots {k-1}^*})
\end{align}
\label{theorem:monotonic}
\end{restatable} where $\mathcal{L}^{k-1} \in  \{\mathcal{L}^{k-1}_{greedy}, \mathcal{L}^{k-1}_{e2e}\}$ and $\mathcal{L}^{k}_{greedy}$ is initialized with the weights $\mathcal{W}^{1^* \ldots {k-1}^*}$. Further details of the GHVAE method and mathematical proofs for these two theorems are in Appendix~\ref{appendix:method} and~\ref{appendix:proof} respectively. 
\begin{table}[t]
\centering
\caption{\small GHVAE vs. Hier-VRNN test performance on CityScapes (mean $\pm$ standard error). All convolutional layers in the 6-module GHVAE are downsized by 40\% to fit into 16GB GPU memory for fair comparison.} 
\vspace{-5px}
\label{tab:vrnn}
\small
\begin{tabular}{|c|c|c|c|c|c|}
\hline
\multirow{2}{*}{Method} & \multirow{2}{*}{FVD $\downarrow$} & \multirow{2}{*}{SSIM $\uparrow$} & \multirow{2}{*}{LPIPS $\downarrow$}  \\
 & & & \\ \hline
GHVAEs & \textbf{418.0$\pm$5.0} & \textbf{74.0$\pm$0.4} & \textbf{0.193$\pm$0.014} \\
Hier-VRNN~\cite{Castrejon_2019_ICCV} & 567.5~\cite{Castrejon_2019_ICCV} & 62.8~\cite{Castrejon_2019_ICCV} & 0.264~\cite{Castrejon_2019_ICCV}\\\hline
\end{tabular}
\vspace{-5px}
\end{table}

%% file: 4-exps.tex
\section{Experimental Evaluation and Analysis}
\label{sec:exp}
We conduct video prediction and real robot experiments to answer six key questions about GHVAEs: \textbf{1)} How do GHVAEs compare to state-of-the-art models in video prediction? 
\textbf{2)} Can GHVAEs achieve monotonic improvement in video prediction accuracy by simply adding more modules, as Theorem~\ref{theorem:monotonic} suggests? \textbf{3)} Does training a GHVAE model end-to-end outperform training greedily per module, as Theorem~\ref{theorem:lowerbound} suggests? \textbf{4)} Does the high expressivity of GHVAEs cause overfitting during training? \textbf{5)} How important is the learned prior network to GHVAEs' performance? \textbf{6)} Does the high expressivity of GHVAEs improve real robot performance? Visualizations and videos are at \url{https://sites.google.com/view/ghvae}, and more qualitative results are in Appendix~\ref{appendix:experiments}. \\

\noindent \textbf{Video Prediction Performance}. To answer the first question, this paper evaluates video prediction methods across five metrics: Fréchet Video Distance (FVD)~\cite{unterthiner2018towards}, Structural Similarity Index Measure (SSIM), Peak Signal-to-noise Ratio (PSNR), Learned Perceptual Image Patch Similarity (LPIPS)~\cite{zhang2018perceptual}, and human preference. FVD and human preference both measure overall visual quality and temporal coherence without reference to the ground truth video. PSNR, SSIM, and LPIPS measure similarity to the ground-truth in different spaces, with LPIPS most accurately representing human perceptual similarity.
To stress-test each method's ability to learn from large and diverse offline video datasets, we use four datasets: RoboNet~\cite{dasari2019robonet} to measure prediction of object interactions, KITTI~\cite{Geiger2013IJRR} and Cityscapes~\cite{Cordts2016Cityscapes} to evaluate the ability to handle partial observability, and Human3.6M~\cite{h36m_pami} to assess prediction of structured motion. This paper compares GHVAEs to SVG'~\cite{svg,villegas2019high} and Hier-VRNN~\cite{Castrejon_2019_ICCV}, which are two state-of-the-art prior methods that use non-hierarchical and hierarchical VAEs respectively. While SAVP~\cite{lee2018stochastic} is another prior method, we empirically found that SAVP underperforms SVG' on these datasets, and therefore omitted SAVP results for simplicity. All metrics are summarized via the mean and standard error over videos in the test set. 

For SVG' in particular, this paper compares to ``SVG' (M=3, K=5)''~\cite{villegas2019high}, which is the largest and best-performing SVG' model that Villegas et al.~\cite{villegas2019high} evaluate and the \textit{largest} version of SVG' that can fit into a 24GB GPU with a batch size of 32. SVG' (M=3, K=5) has 3x larger convolutional LSTMs and 5x larger encoder and decoder convolutional networks compared to the original SVG~\cite{svg} and significantly outperforms the original SVG by 40-60\% in FVD scores~\cite{villegas2019high}. Since Villegas et al.~\cite{villegas2019high} reported the FVD, SSIM, and PSNR performance of ``SVG' (M=3, K=5)'' on KITTI and Human3.6M, we directly compare to their results using the same evaluation methodology. For RoboNet and for evaluating LPIPS and human preference, we re-implement SVG' and report the corresponding performance. In Table~\ref{tab:svg}, the 6-module GHVAE model outperforms SVG' across all three datasets across all metrics. Most saliently, we see a 17-55\% improvement in FVD score and a 13-45\% improvement in LPIPS. Further, we see that humans prefer predictions from the GHVAE model more than $85\%$ of the time. 

To compare to Hier-VRNN~\cite{Castrejon_2019_ICCV}, we use the Cityscapes driving dataset~\cite{Cordts2016Cityscapes}. Since Castrejon et al.~\cite{Castrejon_2019_ICCV} already report FVD, SSIM, and LPIPS performance on Cityscapes, we directly compare against these results using the same evaluation setting. Table~\ref{tab:vrnn} indicates that GHVAEs outperform Hier-VRNN by 26\% in FVD, 18\% in SSIM, and 27\% in LPIPS for Cityscapes when the number of modules reaches six. 

These results indicate that GHVAEs significantly outperform state-of-the-art video prediction models, including hierarchical and non-hierarchical models. The strong performance of GHVAEs mainly originates from the capacity to learn larger models with a stable optimization within the same amount of GPU or TPU memory. For example, even though both GHVAE and SVG' consume 24GB of memory during training, GHVAE contains 599 million parameters while SVG' has 298 million. Next, we perform several ablations to better understand the good performance of GHVAEs.\\
\begin{table}[t]
\centering
\caption{\small Ablation 1: GHVAEs improve monotonically from 2, to 4, and to 6 modules when greedily optimized.}
\vspace{-5px}
\label{tab:ablation1}
\resizebox{1.0\linewidth}{!}{
\begin{tabular}{|c|c|c|c|c|c|c|c|}
\hline
\# of & \multicolumn{4}{c|}{RoboNet Video Prediction Test Performance} \\ 
Modules & FVD $\downarrow$ & PSNR $\uparrow$ & SSIM $\uparrow$ & LPIPS $\downarrow$\\ \hline
6 & \textbf{95.2$\pm$2.6}  & \textbf{24.7$\pm$0.2} &  \textbf{89.1$\pm$0.4} & \textbf{0.036$\pm$0.001} \\
4 & 151.2$\pm$2.3 & 24.2$\pm$0.1& 87.5$\pm$0.4 & 0.059$\pm$0.006\\
2 & 292.4$\pm$11.1 & 23.5$\pm$0.2 & 86.4$\pm$0.2 & 0.106$\pm$0.010 \\\hline
\end{tabular}}
\vspace{-10px}
\end{table}

\noindent \textbf{Ablation 1: Monotonic Improvement and Scalability of GHVAEs}. Given that GHVAEs can be stacked sequentially, it becomes important to determine whether GHVAEs can achieve monotonic improvement by simply adding more GHVAE modules, as suggested by Theorem~\ref{theorem:monotonic}. We observe in Table~\ref{tab:ablation1} that increasing the number of GHVAE modules from 2, to 4, to eventually 6 improves performance across all metrics. These results validate Theorem~\ref{theorem:monotonic} and suggest that greedily adding more modules increases performance monotonically in practice and enables GHVAEs to scale to large datasets.\\

\begin{table}[t]
\centering
\caption{\small Ablation 2: On RoboNet, GHVAEs perform better when optimized greedily than when trained end-to-end.}
\label{tab:ablation2}
\resizebox{\linewidth}{!}{
\begin{tabular}{|c|c|c|c|c|}
\hline
\multirow{2}{*}{Optimization} & \multicolumn{4}{c|}{RoboNet Video Prediction Test Performance} \\ 
& FVD $\downarrow$ & PSNR $\uparrow$ & SSIM $\uparrow$ & LPIPS $\downarrow$\\ \hline
End-to-end Training &  509.9$\pm$6.2 & 21.2$\pm$0.3 & 83.5$\pm$1.0 & 0.148$\pm$0.004\\ \hline
Greedy Training & 95.2$\pm$2.6  & 24.7$\pm$0.2 &  89.1$\pm$0.4 & 0.036$\pm$0.001 \\\hline
  Greedy Training +  & \multirow{2}{*}{\textbf{91.1$\pm$3.1}}& \multirow{2}{*}{\textbf{25.0$\pm$0.2}} & \multirow{2}{*}{\textbf{89.5$\pm$0.5}} & \multirow{2}{*}{\textbf{0.032$\pm$0.003}}\\
  End-to-End Fine-tuning & & &  & \\\hline
\end{tabular}}
\vspace{-10px}
\end{table}
\noindent \textbf{Ablation 2: Greedy vs. End-to-End Optimization of GHVAEs.} End-to-end learning is conventionally preferred over greedy training when GPU or TPU memory constraints are loose. To examine whether this pattern also holds for GHVAEs, we trained a 6-module GHVAE model end-to-end using two 48GB GPUs (since the end-to-end model does not fit in 24GB GPUs) across five separate trials. In addition, we conducted a second experiment in which we fine-tune the greedily trained GHVAE model end-to-end using two 48GB GPUs. We found in Table~\ref{tab:ablation2} that the model was unable to converge to any good performance in any single run compared to the greedy setting. Qualitatively, when optimized end-to-end, GHVAE models need to update each module to improve video prediction quality while preserving the interdependency among individual hidden variables simultaneously, which can lead to optimization difficulties~\cite{sonderby2016train}. Even if GHVAEs can be optimized end-to-end, limited GPU or TPU memory capacity will still make it infeasible to train as the number of modules grows beyond six. However, end-to-end fine-tuning does lead to minor performance gains as indicated by row ``GHVAEs (End-to-End Fine-Tuning, Abl.~2)''. These two experiments imply that greedy training of GHVAEs leads to higher optimization stability than end-to-end training from scratch. They also indicate that end-to-end training of GHVAE can outperform greedy training as suggested by Theorem~\ref{theorem:lowerbound}, so long as the GHVAE model is first pre-trained greedily.\\

\begin{table}[t]
\centering
\caption{\small Ablation 3: Train vs. test performance for a 6-module GHVAE. We observe slight overfitting in all datasets except RoboNet.}
\vspace{-8px}
\label{tab:ablation3}
\resizebox{\linewidth}{!}{
\begin{tabular}{|c|c|c|c|c|c|}
\hline
\multirow{2}{*}{Dataset} & Train & \multicolumn{4}{c|}{Video Prediction Performance} \\ 
& / Test & FVD $\downarrow$ & PSNR $\uparrow$ & SSIM $\uparrow$ & LPIPS $\downarrow$\\ \hline
\multirow{2}{*}{RoboNet} & Train & \textbf{94.4$\pm$3.9} & \textbf{24.9$\pm$0.3} & \textbf{89.3$\pm$0.7} & \textbf{0.036$\pm$0.002}\\
& Test & 95.2$\pm$2.6  & 24.7$\pm$0.2 &  89.1$\pm$0.4 & 0.036$\pm$0.001 \\\hline
\multirow{2}{*}{KITTI} & Train & \textbf{453.5$\pm$12.5} & \textbf{19.4$\pm$0.2} & \textbf{61.4$\pm$1.6} & \textbf{0.209$\pm$0.006}\\
& Test & 552.9$\pm$21.2 & 15.8$\pm$0.1 & 51.2$\pm$2.4 & 0.286$\pm$0.015\\\hline
Human & Train &   \textbf{258.9$\pm$6.8} & \textbf{28.6$\pm$0.3} & \textbf{96.4$\pm$0.1} & \textbf{0.015$\pm$0.002}\\
3.6M & Test & 355.2$\pm$2.9 & 26.7$\pm$0.2 & 94.6$\pm$0.5  & 0.018$\pm$0.002\\\hline
\multirow{2}{*}{Cityscapes} & Train &  \textbf{401.8$\pm$5.4} & \textbf{25.2$\pm$0.1} & \textbf{74.9$\pm$0.1} & \textbf{0.194$\pm$0.006} \\
& Test & \!418.0$\pm$5.0\! & 25.0$\pm$0.1 & 74.0$\pm$0.4 & 0.193$\pm$0.014 \\\hline
\end{tabular}}
\vspace{-5px}
\end{table}
\noindent \textbf{Ablation 3: Train-Test Comparison for GHVAEs.}
Since GHVAEs aim to tackle the underfitting challenge of large-scale video prediction, we now study whether GHVAEs have started to overfit to the training data. We observe in Table~\ref{tab:ablation3} that for RoboNet, a 6-module GHVAE's training performance is similar to its test performance across all four metrics, implying little overfitting. For KITTI, Human3.6M, and Cityscapes, we observe that train performance is better than test performance across most metrics, indicating some overfitting. We hypothesize that this is due to the smaller sizes of these three datasets compared to RoboNet, and, for Human3.6M, because the test set corresponds to two unseen human subjects.\\

\begin{table}[t]
\centering
\caption{\small Ablation 4: Using a learned prior in GHVAEs substantially outperforms a uniform prior particularly in action-conditioned video prediction.}
\vspace{-8px}
\label{tab:ablation4}
\resizebox{\linewidth}{!}{
\begin{tabular}{|c|c|c|c|c|c|}
\hline
\multirow{2}{*}{Dataset} & Learned / &  \multicolumn{4}{c|}{Video Prediction Test Performance} \\ 
 & Uniform & FVD $\downarrow$ & PSNR $\uparrow$ & SSIM $\uparrow$ & LPIPS $\downarrow$\\ \hline
  \multirow{2}{*}{RoboNet} & Learned & \textbf{95.2$\pm$2.6}  & \textbf{24.7$\pm$0.2 }&  \textbf{89.1$\pm$0.4} & \textbf{0.036$\pm$0.001}\\
  & Uniform & 281.4$\pm$1.6 & 22.1$\pm$0.3 & 85.0$\pm$0.4 & 0.58$\pm$0.007\\\hline
  \multirow{2}{*}{KITTI} & Learned & \textbf{552.9$\pm$21.2} & \textbf{15.8$\pm$0.1} & \textbf{51.2$\pm$2.4} & \textbf{0.286$\pm$0.015}\\
 & Uniform & 823.3$\pm$12.0 & 13.0$\pm$0.2 & 46.9$\pm$0.3 & 0.291$\pm$0.005\\ \hline
Human & Learned & \textbf{355.2$\pm$2.9} & \textbf{26.7$\pm$0.2} & \textbf{94.6$\pm$0.5}  & \textbf{0.018$\pm$0.002}\\
3.6M & Uniform & 391.6$\pm$11.1& 26.3$\pm$0.3& 93.0$\pm$0.3& 0.021$\pm$0.002 \\\hline
\multirow{2}{*}{Cityscapes} & Learned & \textbf{418.0$\pm$5.0\!} & \textbf{25.0$\pm$0.1} & \textbf{74.0$\pm$0.4} & \textbf{0.193$\pm$0.014} \\
& Uniform  & 495.2$\pm$1.8 & 24.7$\pm$0.1& 69.1$\pm$0.4 & 0.220$\pm$0.005\\\hline
\end{tabular}}
\vspace{-7px}
\end{table}
\noindent \textbf{Ablation 4: Performance Contribution of Learned Prior.} 
One of GHVAEs' insights is to predict forward the stochastic latent variable only at the deepest layer. Therefore, it may be important to quantify the contribution of the learned prior network to the overall performance. We observe in Table~\ref{tab:ablation4} that using a learned prior significantly outperforms using a uniform diagonal Gaussian prior particularly for action-conditioned datasets. We hypothesize that this is because a learned prior contains information about the action while a uniform prior does not.\\

\begin{figure}[t]
    \centering
    \subfloat[][Train: Random Interaction]{\includegraphics[height=136pt,width=0.45\linewidth]{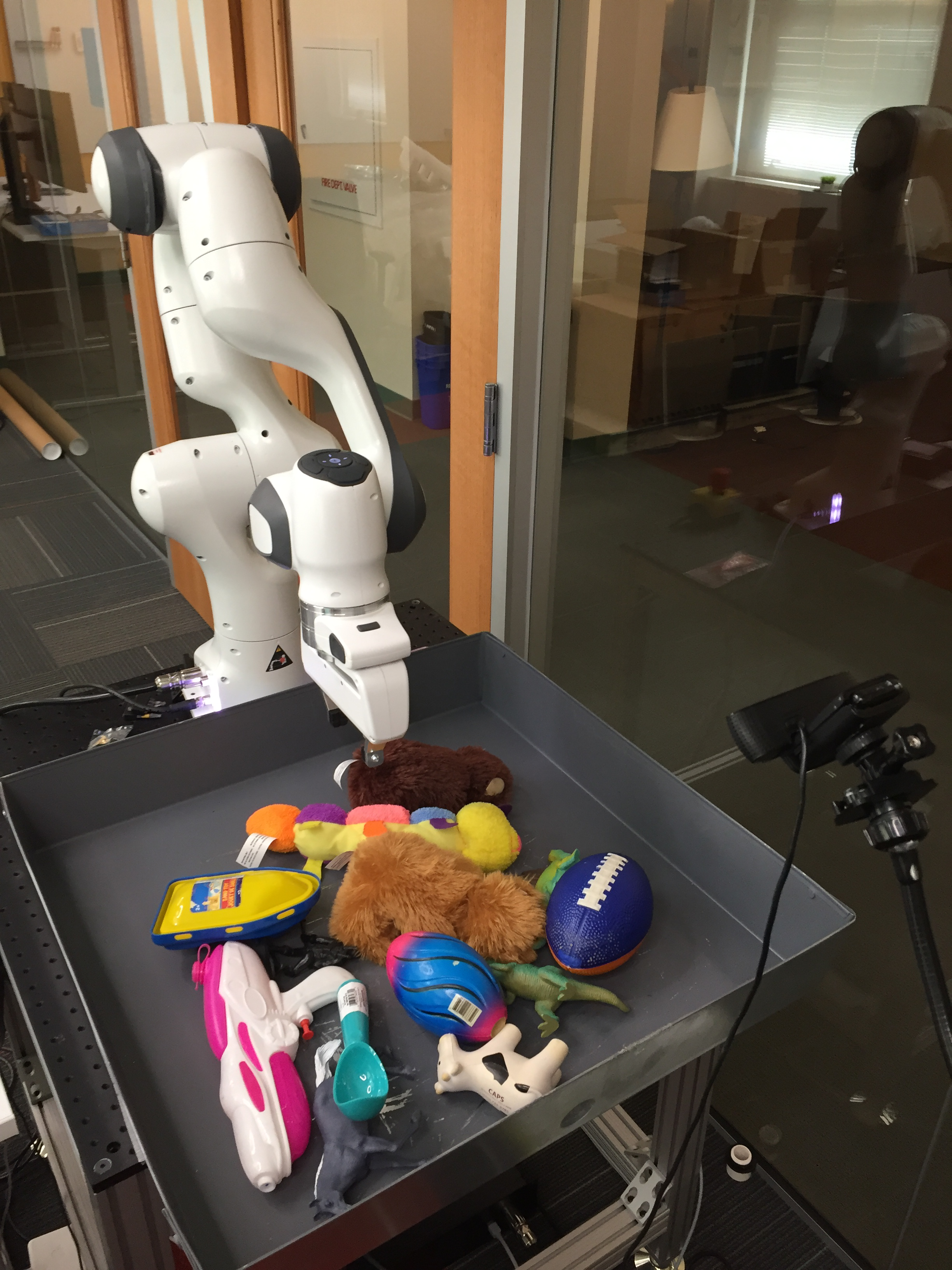}
    \label{robot-train}}\hspace{10pt}
    \subfloat[][Test: Unseen Objects]{
    \includegraphics[height=136pt,width=0.45\linewidth]{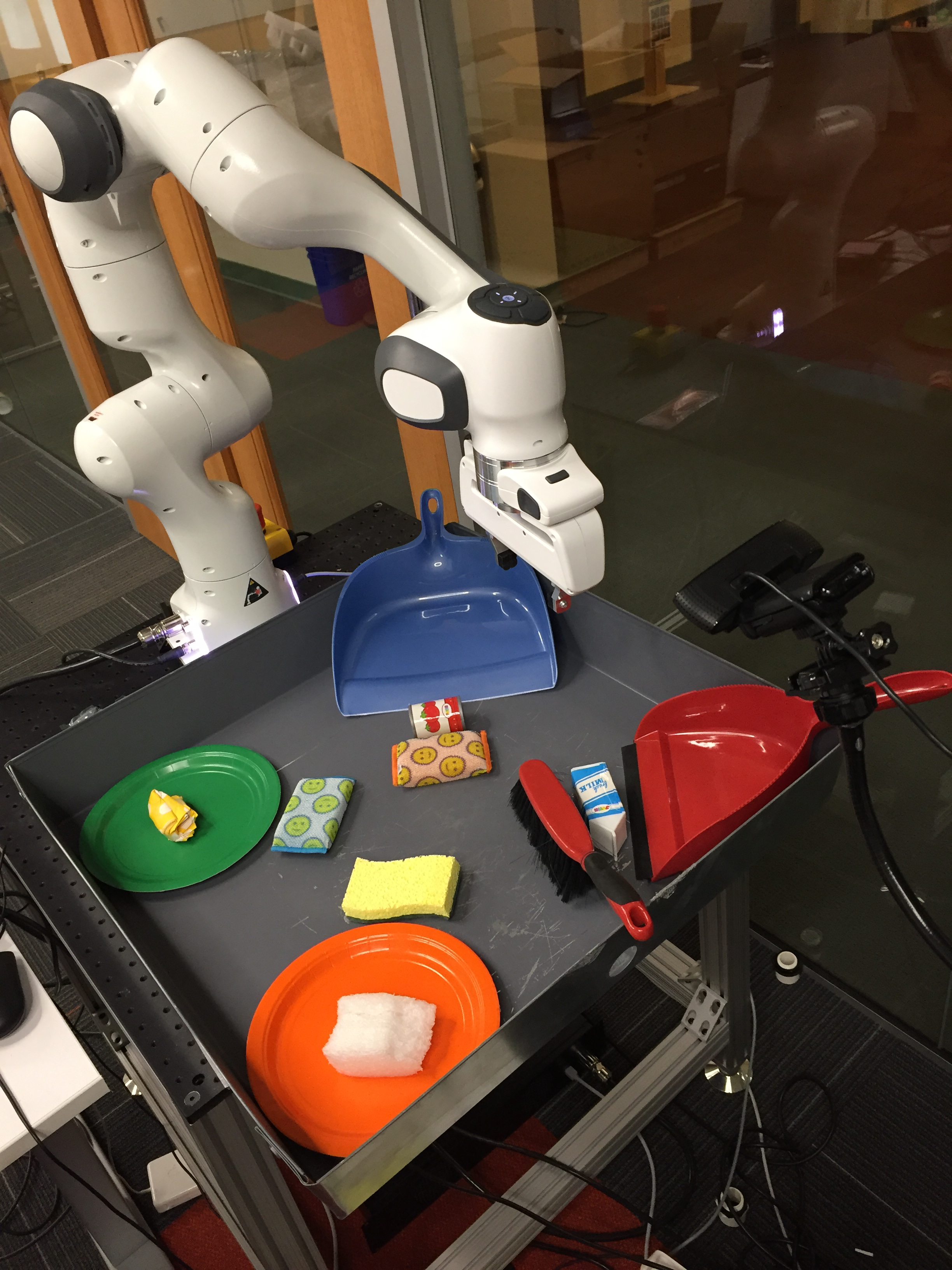}
    \label{robot-test}}
    \vspace{-8px}
    \caption{\small \textbf{Real Robot Experimental Setup.} The Franka robot is equipped with a $45\degree$ black RGB camera. We pre-train each model on RoboNet and fine-tune on an \textit{autonomously} collected dataset of 5000 videos of the robot's random interactions with objects in the bin (Fig.~\ref{robot-train}). Using the trained GHVAE video prediction model, the Franka robot is tested across two tasks: Pick\&Wipe (top and bottom left of bin in Fig.~\ref{robot-test}) and Pick\&Sweep (top and bottom right of bin in Fig.~\ref{robot-test}). All tasks are evaluated on objects, tools, and containers \textit{never seen} during training.}
    \vspace{-2px}
    \label{robot}
\end{figure}

\noindent \textbf{Real Robot Performance}. Finally, we evaluate whether improved video prediction performance translates to greater success on downstream tasks. We consider two manipulation tasks: \textbf{Pick\&Wipe} and \textbf{Pick\&Sweep} on a Franka Emika Panda robot arm. Concretely, each method is given a small, \textit{autonomously collected} training dataset of 5000 videos of random robot interactions with diverse objects such as those in the dark-grey tabletop bin in Fig.~\ref{robot-train}. At test time, to measure generalization, all objects, tools, and containers used are \textit{never seen} during training. Empirically, training directly on this small 5000-video dataset leads to poor generalization to novel objects at test time for all methods. Thus, to enable better generalization, all networks are first pretrained on RoboNet~\cite{dasari2019robonet} and subsequently fine-tuned on this 5000-video dataset. In both tasks, the robot is given a single $64 \times 64$ RGB goal image to indicate the task goal, with no hand-designed rewards provided. The model rollout horizon for each video prediction method is 10, with two prior context frames and a sequence of 10 future actions provided as input. All real-robot results are evaluated across 20 trials. For planning, we perform random shooting (details in Appendix~\ref{appendix:experiments}) with a 4-dimensional action space,
which contains three scalars for the $[x, y, z]$ end-effector translation and one binary scalar for opening vs. closing its parallel-jaw gripper.

In the first \textbf{Pick\&Wipe} task, the robot needs to pick a wiping tool (e.g. sponge, table cloth, etc.) up and wipe all objects off the plate using the wiping tool. The task is successful if the robot picks the wiping tool up and wipe all objects off the plate using the wiping tool within 50 timesteps. In the second \textbf{Pick\&Sweep} task, the robot is required to pick a sweeping tool (e.g. dustpan sweeper, table cloth, or sponge, etc.) up and sweep an object into the dustpan. The task is successful if the target object is swept into the dustpan within 50 timesteps. At the beginning of each task, the wiping or sweeping tool is \textit{not yet} in the robot's gripper, which makes the tasks more difficult. Table~\ref{tab:robot} reveals that a 6-Module GHVAE model outperforms SVG' by $40\%$ and $35\%$ in success rate for Pick\&Wipe and Pick\&Sweep respectively. For Pick\&Wipe, SVG' produces blurry predictions especially when the robot and the plate overlap in the image. This reduces SVG's ability to predict the best action sequence for wiping objects off the plate. In contrast, GHVAE empirically produces accurate predictions of the robot's motion and the position of the wiping tool and the objects. For Pick\&Sweep, SVG' has difficulty predicting the movement of the object during the robot's sweeping motion, leading to more frequent task failures. In contrast, GHVAE predicts plausible robot sweep motions and object movements, reaching an $85\%$ success rate. These results indicate that GHVAEs not only lead to better video prediction performance but that they lead to better downstream performance on real robotic manipulation tasks.
\begin{table}[t]
\centering
\caption{\small GHVAE vs. SVG' real robot performance}
\vspace{-8px}
\label{tab:robot}
\resizebox{0.4\textwidth}{!}{
\begin{tabular}{|c|c|c|}
\hline
\multirow{2}{*}{Method} & \multicolumn{2}{c|}{Test Task Success Rate} \\ 
& Pick\&Wipe Tasks & Pick\&Sweep Tasks \\ \hline
GHVAEs & \textbf{90.0\%} & \textbf{85.0\%} \\
SVG' & 50.0\%  & 50.0\%\\\hline
\end{tabular}}
\vspace{-12px}
\end{table}
\captionsetup[subfloat]{labelformat=empty}

%% file: 5-conclusion.tex
\section{Conclusion}
This paper introduces Greedy Hierarchical VAEs (GHVAEs), which are local VAE modules that can be stacked sequentially and optimized greedily to construct an expressive yet stably optimized hierarchical variational video prediction model. This method significantly outperforms state-of-the-art hierarchical and non-hierarchical video prediction methods by 17-55\% in FVD score across four video datasets and by 35-40\% in real-robot task success rate. Furthermore, GHVAE achieves monotonic improvement by simply stacking more modules. By addressing the underfitting challenge of large-scale video prediction, this work makes it possible for intelligent agents such as robots to learn from large-scale offline video datasets and generalize across a wide range of complex visuomotor tasks through accurate visual foresight. 

While GHVAEs exhibit monotonic improvement, experimenting with GHVAEs beyond six modules is an important direction for future work to better understand the full potential of this method. On the other hand, leveraging this method to enable robotic agents to learn much harder and longer-horizon manipulation and navigation tasks is also an important future direction. Finally, it would be interesting to explore the use of GHVAEs for other generative modeling problems.

%% file: paper.bbl

%% file: robonet.tex
\begin{figure}[t]
  \centering
    \minipage{0.089\linewidth}\textbf{}\endminipage\hfill 
    \minipage{0.089\linewidth}\centering \textbf{$t=1$}\endminipage\hfill 
    \minipage{0.089\linewidth}\centering \textbf{$t=2$}\endminipage\hfill 
    \minipage{0.089\linewidth}\centering \textbf{$t=3$}\endminipage\hfill 
    \minipage{0.089\linewidth}\centering \textbf{$t=4$}\endminipage\hfill 
    \minipage{0.089\linewidth}\centering \textbf{$t=5$}\endminipage\hfill 
    \minipage{0.089\linewidth}\centering \textbf{$t=6$}\endminipage\hfill 
    \minipage{0.089\linewidth}\centering \textbf{$t=7$}\endminipage\hfill 
    \minipage{0.089\linewidth}\centering \textbf{$t=8$}\endminipage\hfill 
    \minipage{0.089\linewidth}\centering \textbf{$t=9$}\endminipage\hfill 
    \minipage{0.089\linewidth}\centering \textbf{$t=10$}\endminipage\hfill 
    \minipage{0.089\linewidth}\textbf{Ground\\Truth (Sawyer)}\endminipage\hfill 
    \minipage{0.089\linewidth}\begin{tikzpicture}\node[inner sep=0pt] (russell) at (0,0){\includegraphics[width=\linewidth]{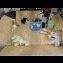}};\draw[red, thick] (-.6,-.2) rectangle (.6,.4);\end{tikzpicture}\endminipage\hfill
    \minipage{0.089\linewidth}\begin{tikzpicture}\node[inner sep=0pt] (russell) at (0,0){\includegraphics[width=\linewidth]{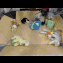}};\draw[red, thick] (-.6,-.2) rectangle (.6,.4);\end{tikzpicture}\endminipage\hfill
    \minipage{0.089\linewidth}\begin{tikzpicture}\node[inner sep=0pt] (russell) at (0,0){\includegraphics[width=\linewidth]{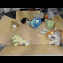}};\draw[red, thick] (-.6,-.2) rectangle (.6,.4);\end{tikzpicture}\endminipage\hfill
    \minipage{0.089\linewidth}\begin{tikzpicture}\node[inner sep=0pt] (russell) at (0,0){\includegraphics[width=\linewidth]{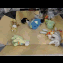}};\draw[red, thick] (-.6,-.2) rectangle (.6,.4);\end{tikzpicture}\endminipage\hfill
    \minipage{0.089\linewidth}\begin{tikzpicture}\node[inner sep=0pt] (russell) at (0,0){\includegraphics[width=\linewidth]{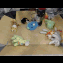}};\draw[red, thick] (-.6,-.2) rectangle (.6,.4);\end{tikzpicture}\endminipage\hfill
    \minipage{0.089\linewidth}\begin{tikzpicture}\node[inner sep=0pt] (russell) at (0,0){\includegraphics[width=\linewidth]{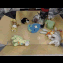}};\draw[red, thick] (-.6,-.2) rectangle (.6,.4);\end{tikzpicture}\endminipage\hfill
    \minipage{0.089\linewidth}\begin{tikzpicture}\node[inner sep=0pt] (russell) at (0,0){\includegraphics[width=\linewidth]{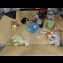}};\draw[red, thick] (-.6,-.2) rectangle (.6,.4);\end{tikzpicture}\endminipage\hfill
    \minipage{0.089\linewidth}\begin{tikzpicture}\node[inner sep=0pt] (russell) at (0,0){\includegraphics[width=\linewidth]{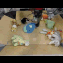}};\draw[red, thick] (-.6,-.2) rectangle (.6,.4);\end{tikzpicture}\endminipage\hfill
    \minipage{0.089\linewidth}\begin{tikzpicture}\node[inner sep=0pt] (russell) at (0,0){\includegraphics[width=\linewidth]{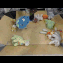}};\draw[red, thick] (-.6,-.2) rectangle (.6,.4);\end{tikzpicture}\endminipage\hfill
    \minipage{0.089\linewidth}\begin{tikzpicture}\node[inner sep=0pt] (russell) at (0,0){\includegraphics[width=\linewidth]{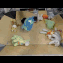}};\draw[red, thick] (-.6,-.2) rectangle (.6,.4);\end{tikzpicture}\endminipage\hfill
    \minipage{0.089\linewidth}\textbf{GHVAE}\endminipage\hfill 
    \minipage{0.089\linewidth}\begin{tikzpicture}\node[inner sep=0pt] (russell) at (0,0){\includegraphics[width=\linewidth]{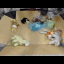}};\draw[red, thick] (-.6,-.2) rectangle (.6,.4);\end{tikzpicture}\endminipage\hfill
    \minipage{0.089\linewidth}\begin{tikzpicture}\node[inner sep=0pt] (russell) at (0,0){\includegraphics[width=\linewidth]{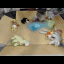}};\draw[red, thick] (-.6,-.2) rectangle (.6,.4);\end{tikzpicture}\endminipage\hfill
    \minipage{0.089\linewidth}\begin{tikzpicture}\node[inner sep=0pt] (russell) at (0,0){\includegraphics[width=\linewidth]{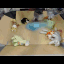}};\draw[red, thick] (-.6,-.2) rectangle (.6,.4);\end{tikzpicture}\endminipage\hfill
    \minipage{0.089\linewidth}\begin{tikzpicture}\node[inner sep=0pt] (russell) at (0,0){\includegraphics[width=\linewidth]{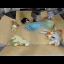}};\draw[red, thick] (-.6,-.2) rectangle (.6,.4);\end{tikzpicture}\endminipage\hfill
    \minipage{0.089\linewidth}\begin{tikzpicture}\node[inner sep=0pt] (russell) at (0,0){\includegraphics[width=\linewidth]{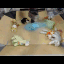}};\draw[red, thick] (-.6,-.2) rectangle (.6,.4);\end{tikzpicture}\endminipage\hfill
    \minipage{0.089\linewidth}\begin{tikzpicture}\node[inner sep=0pt] (russell) at (0,0){\includegraphics[width=\linewidth]{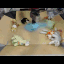}};\draw[red, thick] (-.6,-.2) rectangle (.6,.4);\end{tikzpicture}\endminipage\hfill
    \minipage{0.089\linewidth}\begin{tikzpicture}\node[inner sep=0pt] (russell) at (0,0){\includegraphics[width=\linewidth]{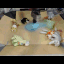}};\draw[red, thick] (-.6,-.2) rectangle (.6,.4);\end{tikzpicture}\endminipage\hfill
    \minipage{0.089\linewidth}\begin{tikzpicture}\node[inner sep=0pt] (russell) at (0,0){\includegraphics[width=\linewidth]{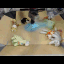}};\draw[red, thick] (-.6,-.2) rectangle (.6,.4);\end{tikzpicture}\endminipage\hfill
    \minipage{0.089\linewidth}\begin{tikzpicture}\node[inner sep=0pt] (russell) at (0,0){\includegraphics[width=\linewidth]{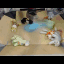}};\draw[red, thick] (-.6,-.2) rectangle (.6,.4);\end{tikzpicture}\endminipage\hfill
    \minipage{0.089\linewidth}\begin{tikzpicture}\node[inner sep=0pt] (russell) at (0,0){\includegraphics[width=\linewidth]{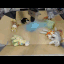}};\draw[red, thick] (-.6,-.2) rectangle (.6,.4);\end{tikzpicture}\endminipage\hfill\minipage{0.089\linewidth}\textbf{SVG' (M=3, K=5)}\endminipage\hfill
    \minipage{0.089\linewidth}\begin{tikzpicture}\node[inner sep=0pt] (russell) at (0,0){\includegraphics[width=\linewidth]{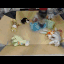}};\draw[red, thick] (-.6,-.2) rectangle (.6,.4);\end{tikzpicture}\endminipage\hfill
    \minipage{0.089\linewidth}\begin{tikzpicture}\node[inner sep=0pt] (russell) at (0,0){\includegraphics[width=\linewidth]{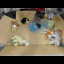}};\draw[red, thick] (-.6,-.2) rectangle (.6,.4);\end{tikzpicture}\endminipage\hfill
    \minipage{0.089\linewidth}\begin{tikzpicture}\node[inner sep=0pt] (russell) at (0,0){\includegraphics[width=\linewidth]{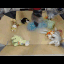}};\draw[red, thick] (-.6,-.2) rectangle (.6,.4);\end{tikzpicture}\endminipage\hfill
    \minipage{0.089\linewidth}\begin{tikzpicture}\node[inner sep=0pt] (russell) at (0,0){\includegraphics[width=\linewidth]{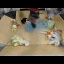}};\draw[red, thick] (-.6,-.2) rectangle (.6,.4);\end{tikzpicture}\endminipage\hfill
    \minipage{0.089\linewidth}\begin{tikzpicture}\node[inner sep=0pt] (russell) at (0,0){\includegraphics[width=\linewidth]{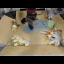}};\draw[red, thick] (-.6,-.2) rectangle (.6,.4);\end{tikzpicture}\endminipage\hfill
    \minipage{0.089\linewidth}\begin{tikzpicture}\node[inner sep=0pt] (russell) at (0,0){\includegraphics[width=\linewidth]{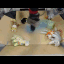}};\draw[red, thick] (-.6,-.2) rectangle (.6,.4);\end{tikzpicture}\endminipage\hfill
    \minipage{0.089\linewidth}\begin{tikzpicture}\node[inner sep=0pt] (russell) at (0,0){\includegraphics[width=\linewidth]{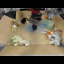}};\draw[red, thick] (-.6,-.2) rectangle (.6,.4);\end{tikzpicture}\endminipage\hfill
    \minipage{0.089\linewidth}\begin{tikzpicture}\node[inner sep=0pt] (russell) at (0,0){\includegraphics[width=\linewidth]{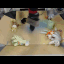}};\draw[red, thick] (-.6,-.2) rectangle (.6,.4);\end{tikzpicture}\endminipage\hfill
    \minipage{0.089\linewidth}\begin{tikzpicture}\node[inner sep=0pt] (russell) at (0,0){\includegraphics[width=\linewidth]{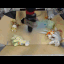}};\draw[red, thick] (-.6,-.2) rectangle (.6,.4);\end{tikzpicture}\endminipage\hfill
    \minipage{0.089\linewidth}\begin{tikzpicture}\node[inner sep=0pt] (russell) at (0,0){\includegraphics[width=\linewidth]{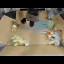}};\draw[red, thick] (-.6,-.2) rectangle (.6,.4);\end{tikzpicture}\endminipage\hfill
    
    \minipage{0.089\linewidth}\textbf{Ground\\Truth (WidowX)}\endminipage\hfill 
    \minipage{0.089\linewidth}\begin{tikzpicture}\node[inner sep=0pt] (russell) at (0,0){\includegraphics[width=\linewidth]{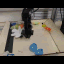}};\draw[red, thick] (-.5,-.3) rectangle (.4,.7);\end{tikzpicture}\endminipage\hfill
    \minipage{0.089\linewidth}\begin{tikzpicture}\node[inner sep=0pt] (russell) at (0,0){\includegraphics[width=\linewidth]{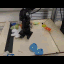}};\draw[red, thick] (-.5,-.3) rectangle (.4,.7);\end{tikzpicture}\endminipage\hfill
    \minipage{0.089\linewidth}\begin{tikzpicture}\node[inner sep=0pt] (russell) at (0,0){\includegraphics[width=\linewidth]{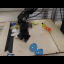}};\draw[red, thick] (-.5,-.3) rectangle (.4,.7);\end{tikzpicture}\endminipage\hfill
    \minipage{0.089\linewidth}\begin{tikzpicture}\node[inner sep=0pt] (russell) at (0,0){\includegraphics[width=\linewidth]{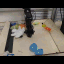}};\draw[red, thick] (-.5,-.3) rectangle (.4,.7);\end{tikzpicture}\endminipage\hfill
    \minipage{0.089\linewidth}\begin{tikzpicture}\node[inner sep=0pt] (russell) at (0,0){\includegraphics[width=\linewidth]{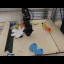}};\draw[red, thick] (-.5,-.3) rectangle (.4,.7);\end{tikzpicture}\endminipage\hfill
    \minipage{0.089\linewidth}\begin{tikzpicture}\node[inner sep=0pt] (russell) at (0,0){\includegraphics[width=\linewidth]{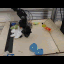}};\draw[red, thick] (-.5,-.3) rectangle (.4,.7);\end{tikzpicture}\endminipage\hfill
    \minipage{0.089\linewidth}\begin{tikzpicture}\node[inner sep=0pt] (russell) at (0,0){\includegraphics[width=\linewidth]{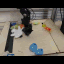}};\draw[red, thick] (-.5,-.3) rectangle (.4,.7);\end{tikzpicture}\endminipage\hfill
    \minipage{0.089\linewidth}\begin{tikzpicture}\node[inner sep=0pt] (russell) at (0,0){\includegraphics[width=\linewidth]{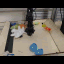}};\draw[red, thick] (-.5,-.3) rectangle (.4,.7);\end{tikzpicture}\endminipage\hfill
    \minipage{0.089\linewidth}\begin{tikzpicture}\node[inner sep=0pt] (russell) at (0,0){\includegraphics[width=\linewidth]{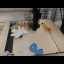}};\draw[red, thick] (-.5,-.3) rectangle (.4,.7);\end{tikzpicture}\endminipage\hfill
    \minipage{0.089\linewidth}\begin{tikzpicture}\node[inner sep=0pt] (russell) at (0,0){\includegraphics[width=\linewidth]{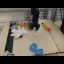}};\draw[red, thick] (-.5,-.3) rectangle (.4,.7);\end{tikzpicture}\endminipage\hfill
    \minipage{0.089\linewidth}\textbf{GHVAE}\endminipage\hfill 
    \minipage{0.089\linewidth}\begin{tikzpicture}\node[inner sep=0pt] (russell) at (0,0){\includegraphics[width=\linewidth]{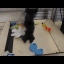}};\draw[red, thick] (-.5,-.3) rectangle (.4,.7);\end{tikzpicture}\endminipage\hfill
    \minipage{0.089\linewidth}\begin{tikzpicture}\node[inner sep=0pt] (russell) at (0,0){\includegraphics[width=\linewidth]{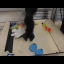}};\draw[red, thick] (-.5,-.3) rectangle (.4,.7);\end{tikzpicture}\endminipage\hfill
    \minipage{0.089\linewidth}\begin{tikzpicture}\node[inner sep=0pt] (russell) at (0,0){\includegraphics[width=\linewidth]{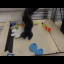}};\draw[red, thick] (-.5,-.3) rectangle (.4,.7);\end{tikzpicture}\endminipage\hfill
    \minipage{0.089\linewidth}\begin{tikzpicture}\node[inner sep=0pt] (russell) at (0,0){\includegraphics[width=\linewidth]{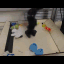}};\draw[red, thick] (-.5,-.3) rectangle (.4,.7);\end{tikzpicture}\endminipage\hfill
    \minipage{0.089\linewidth}\begin{tikzpicture}\node[inner sep=0pt] (russell) at (0,0){\includegraphics[width=\linewidth]{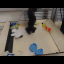}};\draw[red, thick] (-.5,-.3) rectangle (.4,.7);\end{tikzpicture}\endminipage\hfill
    \minipage{0.089\linewidth}\begin{tikzpicture}\node[inner sep=0pt] (russell) at (0,0){\includegraphics[width=\linewidth]{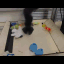}};\draw[red, thick] (-.5,-.3) rectangle (.4,.7);\end{tikzpicture}\endminipage\hfill
    \minipage{0.089\linewidth}\begin{tikzpicture}\node[inner sep=0pt] (russell) at (0,0){\includegraphics[width=\linewidth]{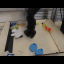}};\draw[red, thick] (-.5,-.3) rectangle (.4,.7);\end{tikzpicture}\endminipage\hfill
    \minipage{0.089\linewidth}\begin{tikzpicture}\node[inner sep=0pt] (russell) at (0,0){\includegraphics[width=\linewidth]{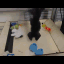}};\draw[red, thick] (-.5,-.3) rectangle (.4,.7);\end{tikzpicture}\endminipage\hfill
    \minipage{0.089\linewidth}\begin{tikzpicture}\node[inner sep=0pt] (russell) at (0,0){\includegraphics[width=\linewidth]{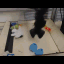}};\draw[red, thick] (-.5,-.3) rectangle (.4,.7);\end{tikzpicture}\endminipage\hfill
    \minipage{0.089\linewidth}\begin{tikzpicture}\node[inner sep=0pt] (russell) at (0,0){\includegraphics[width=\linewidth]{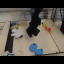}};\draw[red, thick] (-.5,-.3) rectangle (.4,.7);\end{tikzpicture}\endminipage\hfill\minipage{0.089\linewidth}\textbf{SVG' (M=3, K=5)}\endminipage\hfill
    \minipage{0.089\linewidth}\begin{tikzpicture}\node[inner sep=0pt] (russell) at (0,0){\includegraphics[width=\linewidth]{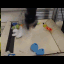}};\draw[red, thick] (-.5,-.3) rectangle (.4,.7);\end{tikzpicture}\endminipage\hfill
    \minipage{0.089\linewidth}\begin{tikzpicture}\node[inner sep=0pt] (russell) at (0,0){\includegraphics[width=\linewidth]{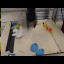}};\draw[red, thick] (-.5,-.3) rectangle (.4,.7);\end{tikzpicture}\endminipage\hfill
    \minipage{0.089\linewidth}\begin{tikzpicture}\node[inner sep=0pt] (russell) at (0,0){\includegraphics[width=\linewidth]{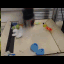}};\draw[red, thick] (-.5,-.3) rectangle (.4,.7);\end{tikzpicture}\endminipage\hfill
    \minipage{0.089\linewidth}\begin{tikzpicture}\node[inner sep=0pt] (russell) at (0,0){\includegraphics[width=\linewidth]{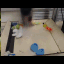}};\draw[red, thick] (-.5,-.3) rectangle (.4,.7);\end{tikzpicture}\endminipage\hfill
    \minipage{0.089\linewidth}\begin{tikzpicture}\node[inner sep=0pt] (russell) at (0,0){\includegraphics[width=\linewidth]{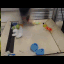}};\draw[red, thick] (-.5,-.3) rectangle (.4,.7);\end{tikzpicture}\endminipage\hfill
    \minipage{0.089\linewidth}\begin{tikzpicture}\node[inner sep=0pt] (russell) at (0,0){\includegraphics[width=\linewidth]{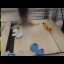}};\draw[red, thick] (-.5,-.3) rectangle (.4,.7);\end{tikzpicture}\endminipage\hfill
    \minipage{0.089\linewidth}\begin{tikzpicture}\node[inner sep=0pt] (russell) at (0,0){\includegraphics[width=\linewidth]{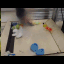}};\draw[red, thick] (-.5,-.3) rectangle (.4,.7);\end{tikzpicture}\endminipage\hfill
    \minipage{0.089\linewidth}\begin{tikzpicture}\node[inner sep=0pt] (russell) at (0,0){\includegraphics[width=\linewidth]{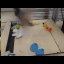}};\draw[red, thick] (-.5,-.3) rectangle (.4,.7);\end{tikzpicture}\endminipage\hfill
    \minipage{0.089\linewidth}\begin{tikzpicture}\node[inner sep=0pt] (russell) at (0,0){\includegraphics[width=\linewidth]{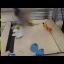}};\draw[red, thick] (-.5,-.3) rectangle (.4,.7);\end{tikzpicture}\endminipage\hfill
    \minipage{0.089\linewidth}\begin{tikzpicture}\node[inner sep=0pt] (russell) at (0,0){\includegraphics[width=\linewidth]{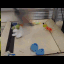}};\draw[red, thick] (-.5,-.3) rectangle (.4,.7);\end{tikzpicture}\endminipage\hfill
    
    \minipage{0.089\linewidth}\textbf{Ground\\Truth (Franka)}\endminipage\hfill 
    \minipage{0.089\linewidth}\begin{tikzpicture}\node[inner sep=0pt] (russell) at (0,0){\includegraphics[width=\linewidth]{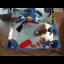}};\draw[red, thick] (-.1,-.1) rectangle (.7,.7);\end{tikzpicture}\endminipage\hfill
    \minipage{0.089\linewidth}\begin{tikzpicture}\node[inner sep=0pt] (russell) at (0,0){\includegraphics[width=\linewidth]{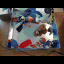}};\draw[red, thick] (-.1,-.1) rectangle (.7,.7);\end{tikzpicture}\endminipage\hfill
    \minipage{0.089\linewidth}\begin{tikzpicture}\node[inner sep=0pt] (russell) at (0,0){\includegraphics[width=\linewidth]{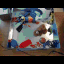}};\draw[red, thick] (-.1,-.1) rectangle (.7,.7);\end{tikzpicture}\endminipage\hfill
    \minipage{0.089\linewidth}\begin{tikzpicture}\node[inner sep=0pt] (russell) at (0,0){\includegraphics[width=\linewidth]{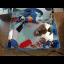}};\draw[red, thick] (-.1,-.1) rectangle (.7,.7);\end{tikzpicture}\endminipage\hfill
    \minipage{0.089\linewidth}\begin{tikzpicture}\node[inner sep=0pt] (russell) at (0,0){\includegraphics[width=\linewidth]{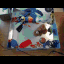}};\draw[red, thick] (-.1,-.1) rectangle (.7,.7);\end{tikzpicture}\endminipage\hfill
    \minipage{0.089\linewidth}\begin{tikzpicture}\node[inner sep=0pt] (russell) at (0,0){\includegraphics[width=\linewidth]{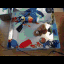}};\draw[red, thick] (-.1,-.1) rectangle (.7,.7);\end{tikzpicture}\endminipage\hfill
    \minipage{0.089\linewidth}\begin{tikzpicture}\node[inner sep=0pt] (russell) at (0,0){\includegraphics[width=\linewidth]{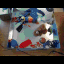}};\draw[red, thick] (-.1,-.1) rectangle (.7,.7);\end{tikzpicture}\endminipage\hfill
    \minipage{0.089\linewidth}\begin{tikzpicture}\node[inner sep=0pt] (russell) at (0,0){\includegraphics[width=\linewidth]{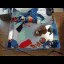}};\draw[red, thick] (-.1,-.1) rectangle (.7,.7);\end{tikzpicture}\endminipage\hfill
    \minipage{0.089\linewidth}\begin{tikzpicture}\node[inner sep=0pt] (russell) at (0,0){\includegraphics[width=\linewidth]{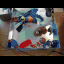}};\draw[red, thick] (-.1,-.1) rectangle (.7,.7);\end{tikzpicture}\endminipage\hfill
    \minipage{0.089\linewidth}\begin{tikzpicture}\node[inner sep=0pt] (russell) at (0,0){\includegraphics[width=\linewidth]{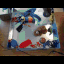}};\draw[red, thick] (-.1,-.1) rectangle (.7,.7);\end{tikzpicture}\endminipage\hfill
    \minipage{0.089\linewidth}\textbf{GHVAE}\endminipage\hfill 
    \minipage{0.089\linewidth}\begin{tikzpicture}\node[inner sep=0pt] (russell) at (0,0){\includegraphics[width=\linewidth]{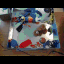}};\draw[red, thick] (-.1,-.1) rectangle (.7,.7);\end{tikzpicture}\endminipage\hfill
    \minipage{0.089\linewidth}\begin{tikzpicture}\node[inner sep=0pt] (russell) at (0,0){\includegraphics[width=\linewidth]{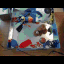}};\draw[red, thick] (-.1,-.1) rectangle (.7,.7);\end{tikzpicture}\endminipage\hfill
    \minipage{0.089\linewidth}\begin{tikzpicture}\node[inner sep=0pt] (russell) at (0,0){\includegraphics[width=\linewidth]{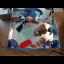}};\draw[red, thick] (-.1,-.1) rectangle (.7,.7);\end{tikzpicture}\endminipage\hfill
    \minipage{0.089\linewidth}\begin{tikzpicture}\node[inner sep=0pt] (russell) at (0,0){\includegraphics[width=\linewidth]{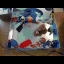}};\draw[red, thick] (-.1,-.1) rectangle (.7,.7);\end{tikzpicture}\endminipage\hfill
    \minipage{0.089\linewidth}\begin{tikzpicture}\node[inner sep=0pt] (russell) at (0,0){\includegraphics[width=\linewidth]{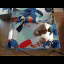}};\draw[red, thick] (-.1,-.1) rectangle (.7,.7);\end{tikzpicture}\endminipage\hfill
    \minipage{0.089\linewidth}\begin{tikzpicture}\node[inner sep=0pt] (russell) at (0,0){\includegraphics[width=\linewidth]{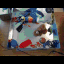}};\draw[red, thick] (-.1,-.1) rectangle (.7,.7);\end{tikzpicture}\endminipage\hfill
    \minipage{0.089\linewidth}\begin{tikzpicture}\node[inner sep=0pt] (russell) at (0,0){\includegraphics[width=\linewidth]{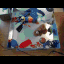}};\draw[red, thick] (-.1,-.1) rectangle (.7,.7);\end{tikzpicture}\endminipage\hfill
    \minipage{0.089\linewidth}\begin{tikzpicture}\node[inner sep=0pt] (russell) at (0,0){\includegraphics[width=\linewidth]{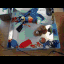}};\draw[red, thick] (-.1,-.1) rectangle (.7,.7);\end{tikzpicture}\endminipage\hfill
    \minipage{0.089\linewidth}\begin{tikzpicture}\node[inner sep=0pt] (russell) at (0,0){\includegraphics[width=\linewidth]{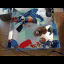}};\draw[red, thick] (-.1,-.1) rectangle (.7,.7);\end{tikzpicture}\endminipage\hfill
    \minipage{0.089\linewidth}\begin{tikzpicture}\node[inner sep=0pt] (russell) at (0,0){\includegraphics[width=\linewidth]{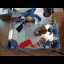}};\draw[red, thick] (-.1,-.1) rectangle (.7,.7);\end{tikzpicture}\endminipage\hfill
    \minipage{0.089\linewidth}\textbf{SVG' (M=3, K=5)}\endminipage\hfill
    \minipage{0.089\linewidth}\begin{tikzpicture}\node[inner sep=0pt] (russell) at (0,0){\includegraphics[width=\linewidth]{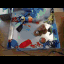}};\draw[red, thick] (-.1,-.1) rectangle (.7,.7);\end{tikzpicture}\endminipage\hfill
    \minipage{0.089\linewidth}\begin{tikzpicture}\node[inner sep=0pt] (russell) at (0,0){\includegraphics[width=\linewidth]{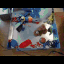}};\draw[red, thick] (-.1,-.1) rectangle (.7,.7);\end{tikzpicture}\endminipage\hfill
    \minipage{0.089\linewidth}\begin{tikzpicture}\node[inner sep=0pt] (russell) at (0,0){\includegraphics[width=\linewidth]{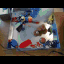}};\draw[red, thick] (-.1,-.1) rectangle (.7,.7);\end{tikzpicture}\endminipage\hfill
    \minipage{0.089\linewidth}\begin{tikzpicture}\node[inner sep=0pt] (russell) at (0,0){\includegraphics[width=\linewidth]{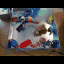}};\draw[red, thick] (-.1,-.1) rectangle (.7,.7);\end{tikzpicture}\endminipage\hfill
    \minipage{0.089\linewidth}\begin{tikzpicture}\node[inner sep=0pt] (russell) at (0,0){\includegraphics[width=\linewidth]{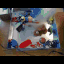}};\draw[red, thick] (-.1,-.1) rectangle (.7,.7);\end{tikzpicture}\endminipage\hfill
    \minipage{0.089\linewidth}\begin{tikzpicture}\node[inner sep=0pt] (russell) at (0,0){\includegraphics[width=\linewidth]{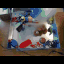}};\draw[red, thick] (-.1,-.1) rectangle (.7,.7);\end{tikzpicture}\endminipage\hfill
    \minipage{0.089\linewidth}\begin{tikzpicture}\node[inner sep=0pt] (russell) at (0,0){\includegraphics[width=\linewidth]{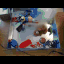}};\draw[red, thick] (-.1,-.1) rectangle (.7,.7);\end{tikzpicture}\endminipage\hfill
    \minipage{0.089\linewidth}\begin{tikzpicture}\node[inner sep=0pt] (russell) at (0,0){\includegraphics[width=\linewidth]{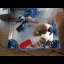}};\draw[red, thick] (-.1,-.1) rectangle (.7,.7);\end{tikzpicture}\endminipage\hfill
    \minipage{0.089\linewidth}\begin{tikzpicture}\node[inner sep=0pt] (russell) at (0,0){\includegraphics[width=\linewidth]{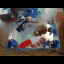}};\draw[red, thick] (-.1,-.1) rectangle (.7,.7);\end{tikzpicture}\endminipage\hfill
    \minipage{0.089\linewidth}\begin{tikzpicture}\node[inner sep=0pt] (russell) at (0,0){\includegraphics[width=\linewidth]{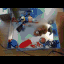}};\draw[red, thick] (-.1,-.1) rectangle (.7,.7);\end{tikzpicture}\endminipage\hfill
    \minipage{0.089\linewidth}\textbf{Ground\\Truth (Baxter)}\endminipage\hfill 
    \minipage{0.089\linewidth}\begin{tikzpicture}\node[inner sep=0pt] (russell) at (0,0){\includegraphics[width=\linewidth]{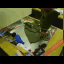}};\draw[red, thick] (-.1,-.4) rectangle (.7,.7);\end{tikzpicture}\endminipage\hfill
    \minipage{0.089\linewidth}\begin{tikzpicture}\node[inner sep=0pt] (russell) at (0,0){\includegraphics[width=\linewidth]{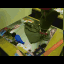}};\draw[red, thick] (-.1,-.4) rectangle (.7,.7);\end{tikzpicture}\endminipage\hfill
    \minipage{0.089\linewidth}\begin{tikzpicture}\node[inner sep=0pt] (russell) at (0,0){\includegraphics[width=\linewidth]{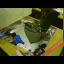}};\draw[red, thick] (-.1,-.4) rectangle (.7,.7);\end{tikzpicture}\endminipage\hfill
    \minipage{0.089\linewidth}\begin{tikzpicture}\node[inner sep=0pt] (russell) at (0,0){\includegraphics[width=\linewidth]{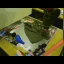}};\draw[red, thick] (-.1,-.4) rectangle (.7,.7);\end{tikzpicture}\endminipage\hfill
    \minipage{0.089\linewidth}\begin{tikzpicture}\node[inner sep=0pt] (russell) at (0,0){\includegraphics[width=\linewidth]{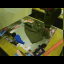}};\draw[red, thick] (-.1,-.4) rectangle (.7,.7);\end{tikzpicture}\endminipage\hfill
    \minipage{0.089\linewidth}\begin{tikzpicture}\node[inner sep=0pt] (russell) at (0,0){\includegraphics[width=\linewidth]{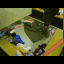}};\draw[red, thick] (-.1,-.4) rectangle (.7,.7);\end{tikzpicture}\endminipage\hfill
    \minipage{0.089\linewidth}\begin{tikzpicture}\node[inner sep=0pt] (russell) at (0,0){\includegraphics[width=\linewidth]{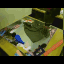}};\draw[red, thick] (-.1,-.4) rectangle (.7,.7);\end{tikzpicture}\endminipage\hfill
    \minipage{0.089\linewidth}\begin{tikzpicture}\node[inner sep=0pt] (russell) at (0,0){\includegraphics[width=\linewidth]{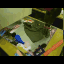}};\draw[red, thick] (-.1,-.4) rectangle (.7,.7);\end{tikzpicture}\endminipage\hfill
    \minipage{0.089\linewidth}\begin{tikzpicture}\node[inner sep=0pt] (russell) at (0,0){\includegraphics[width=\linewidth]{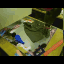}};\draw[red, thick] (-.1,-.4) rectangle (.7,.7);\end{tikzpicture}\endminipage\hfill
    \minipage{0.089\linewidth}\begin{tikzpicture}\node[inner sep=0pt] (russell) at (0,0){\includegraphics[width=\linewidth]{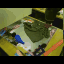}};\draw[red, thick] (-.1,-.4) rectangle (.7,.7);\end{tikzpicture}\endminipage\hfill
    \minipage{0.089\linewidth}\textbf{GHVAE}\endminipage\hfill 
    \minipage{0.089\linewidth}\begin{tikzpicture}\node[inner sep=0pt] (russell) at (0,0){\includegraphics[width=\linewidth]{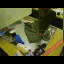}};\draw[red, thick] (-.1,-.4) rectangle (.7,.7);\end{tikzpicture}\endminipage\hfill
    \minipage{0.089\linewidth}\begin{tikzpicture}\node[inner sep=0pt] (russell) at (0,0){\includegraphics[width=\linewidth]{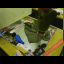}};\draw[red, thick] (-.1,-.4) rectangle (.7,.7);\end{tikzpicture}\endminipage\hfill
    \minipage{0.089\linewidth}\begin{tikzpicture}\node[inner sep=0pt] (russell) at (0,0){\includegraphics[width=\linewidth]{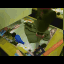}};\draw[red, thick] (-.1,-.4) rectangle (.7,.7);\end{tikzpicture}\endminipage\hfill
    \minipage{0.089\linewidth}\begin{tikzpicture}\node[inner sep=0pt] (russell) at (0,0){\includegraphics[width=\linewidth]{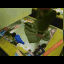}};\draw[red, thick] (-.1,-.4) rectangle (.7,.7);\end{tikzpicture}\endminipage\hfill
    \minipage{0.089\linewidth}\begin{tikzpicture}\node[inner sep=0pt] (russell) at (0,0){\includegraphics[width=\linewidth]{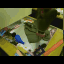}};\draw[red, thick] (-.1,-.4) rectangle (.7,.7);\end{tikzpicture}\endminipage\hfill
    \minipage{0.089\linewidth}\begin{tikzpicture}\node[inner sep=0pt] (russell) at (0,0){\includegraphics[width=\linewidth]{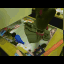}};\draw[red, thick] (-.1,-.4) rectangle (.7,.7);\end{tikzpicture}\endminipage\hfill
    \minipage{0.089\linewidth}\begin{tikzpicture}\node[inner sep=0pt] (russell) at (0,0){\includegraphics[width=\linewidth]{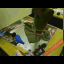}};\draw[red, thick] (-.1,-.4) rectangle (.7,.7);\end{tikzpicture}\endminipage\hfill
    \minipage{0.089\linewidth}\begin{tikzpicture}\node[inner sep=0pt] (russell) at (0,0){\includegraphics[width=\linewidth]{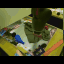}};\draw[red, thick] (-.1,-.4) rectangle (.7,.7);\end{tikzpicture}\endminipage\hfill
    \minipage{0.089\linewidth}\begin{tikzpicture}\node[inner sep=0pt] (russell) at (0,0){\includegraphics[width=\linewidth]{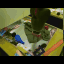}};\draw[red, thick] (-.1,-.4) rectangle (.7,.7);\end{tikzpicture}\endminipage\hfill
    \minipage{0.089\linewidth}\begin{tikzpicture}\node[inner sep=0pt] (russell) at (0,0){\includegraphics[width=\linewidth]{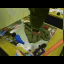}};\draw[red, thick] (-.1,-.4) rectangle (.7,.7);\end{tikzpicture}\endminipage\hfill\minipage{0.089\linewidth}\textbf{SVG' (M=3, K=5)}\endminipage\hfill
    \minipage{0.089\linewidth}\begin{tikzpicture}\node[inner sep=0pt] (russell) at (0,0){\includegraphics[width=\linewidth]{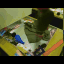}};\draw[red, thick] (-.1,-.4) rectangle (.7,.7);\end{tikzpicture}\endminipage\hfill
    \minipage{0.089\linewidth}\begin{tikzpicture}\node[inner sep=0pt] (russell) at (0,0){\includegraphics[width=\linewidth]{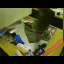}};\draw[red, thick] (-.1,-.4) rectangle (.7,.7);\end{tikzpicture}\endminipage\hfill
    \minipage{0.089\linewidth}\begin{tikzpicture}\node[inner sep=0pt] (russell) at (0,0){\includegraphics[width=\linewidth]{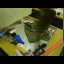}};\draw[red, thick] (-.1,-.4) rectangle (.7,.7);\end{tikzpicture}\endminipage\hfill
    \minipage{0.089\linewidth}\begin{tikzpicture}\node[inner sep=0pt] (russell) at (0,0){\includegraphics[width=\linewidth]{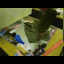}};\draw[red, thick] (-.1,-.4) rectangle (.7,.7);\end{tikzpicture}\endminipage\hfill
    \minipage{0.089\linewidth}\begin{tikzpicture}\node[inner sep=0pt] (russell) at (0,0){\includegraphics[width=\linewidth]{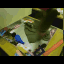}};\draw[red, thick] (-.1,-.4) rectangle (.7,.7);\end{tikzpicture}\endminipage\hfill
    \minipage{0.089\linewidth}\begin{tikzpicture}\node[inner sep=0pt] (russell) at (0,0){\includegraphics[width=\linewidth]{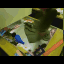}};\draw[red, thick] (-.1,-.4) rectangle (.7,.7);\end{tikzpicture}\endminipage\hfill
    \minipage{0.089\linewidth}\begin{tikzpicture}\node[inner sep=0pt] (russell) at (0,0){\includegraphics[width=\linewidth]{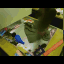}};\draw[red, thick] (-.1,-.4) rectangle (.7,.7);\end{tikzpicture}\endminipage\hfill
    \minipage{0.089\linewidth}\begin{tikzpicture}\node[inner sep=0pt] (russell) at (0,0){\includegraphics[width=\linewidth]{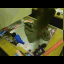}};\draw[red, thick] (-.1,-.4) rectangle (.7,.7);\end{tikzpicture}\endminipage\hfill
    \minipage{0.089\linewidth}\begin{tikzpicture}\node[inner sep=0pt] (russell) at (0,0){\includegraphics[width=\linewidth]{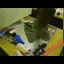}};\draw[red, thick] (-.1,-.4) rectangle (.7,.7);\end{tikzpicture}\endminipage\hfill
    \minipage{0.089\linewidth}\begin{tikzpicture}\node[inner sep=0pt] (russell) at (0,0){\includegraphics[width=\linewidth]{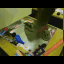}};\draw[red, thick] (-.1,-.4) rectangle (.7,.7);\end{tikzpicture}\endminipage\hfill
  \caption{RoboNet Video Prediction. Specifically, we provide examples for various physical robots in RoboNet: Sawyer, WidowX, Franka, and Baxter. Both GHVAE and SVG' (M=3, K=5) are given the same two context images. Here, a 6-module GHVAE model exhibits visible performance superiority over SVG' (M=3, K=5) on generating realistic object (Sawyer) and robot movements (WidowX, Franka, Baxter). \color{red}{The red boxes} highlight the differences.}\label{robonet}
\end{figure}

%% file: kitti.tex
\begin{figure}[t]
  \centering
    \minipage{0.089\linewidth}\textbf{Ground\\Truth}\endminipage\hfill 
    \minipage{0.089\linewidth}\includegraphics[width=\linewidth]{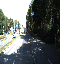}{}\endminipage\hfill
    \minipage{0.089\linewidth}\includegraphics[width=\linewidth]{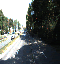}{}\endminipage\hfill
    \minipage{0.089\linewidth}\includegraphics[width=\linewidth]{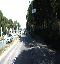}{}\endminipage\hfill
    \minipage{0.089\linewidth}\includegraphics[width=\linewidth]{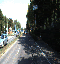}{}\endminipage\hfill
    \minipage{0.089\linewidth}\includegraphics[width=\linewidth]{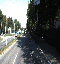}{}\endminipage\hfill
    \minipage{0.089\linewidth}\includegraphics[width=\linewidth]{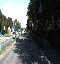}{}\endminipage\hfill
    \minipage{0.089\linewidth}\includegraphics[width=\linewidth]{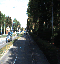}{}\endminipage\hfill
    \minipage{0.089\linewidth}\includegraphics[width=\linewidth]{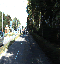}{}\endminipage\hfill
    \minipage{0.089\linewidth}\includegraphics[width=\linewidth]{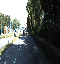}{}\endminipage\hfill
    \minipage{0.089\linewidth}\includegraphics[width=\linewidth]{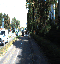}{}\endminipage\hfill
    \minipage{0.089\linewidth}\textbf{GHVAE}\endminipage\hfill 
    \minipage{0.089\linewidth}\includegraphics[width=\linewidth]{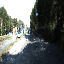}{}\endminipage\hfill
    \minipage{0.089\linewidth}\includegraphics[width=\linewidth]{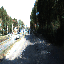}{}\endminipage\hfill
    \minipage{0.089\linewidth}\includegraphics[width=\linewidth]{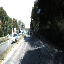}{}\endminipage\hfill
    \minipage{0.089\linewidth}\includegraphics[width=\linewidth]{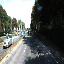}{}\endminipage\hfill
    \minipage{0.089\linewidth}\includegraphics[width=\linewidth]{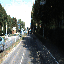}{}\endminipage\hfill
    \minipage{0.089\linewidth}\includegraphics[width=\linewidth]{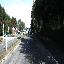}{}\endminipage\hfill
    \minipage{0.089\linewidth}\includegraphics[width=\linewidth]{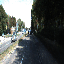}{}\endminipage\hfill
    \minipage{0.089\linewidth}\includegraphics[width=\linewidth]{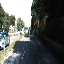}{}\endminipage\hfill
    \minipage{0.089\linewidth}\includegraphics[width=\linewidth]{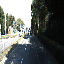}{}\endminipage\hfill
    \minipage{0.089\linewidth}\includegraphics[width=\linewidth]{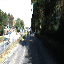}{}\endminipage\hfill
    \minipage{0.089\linewidth}\textbf{SVG' (M=3, K=5)}\endminipage\hfill 
    \minipage{0.089\linewidth}\includegraphics[width=\linewidth]{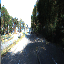}{\caption*{$t=1$}}\endminipage\hfill
    \minipage{0.089\linewidth}\includegraphics[width=\linewidth]{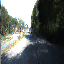}{\caption*{$t=2$}}\endminipage\hfill
    \minipage{0.089\linewidth}\includegraphics[width=\linewidth]{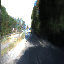}{\caption*{$t=4$}}\endminipage\hfill
    \minipage{0.089\linewidth}\includegraphics[width=\linewidth]{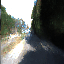}{\caption*{$t=6$}}\endminipage\hfill
    \minipage{0.089\linewidth}\includegraphics[width=\linewidth]{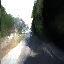}{\caption*{$t=8$}}\endminipage\hfill
    \minipage{0.089\linewidth}\includegraphics[width=\linewidth]{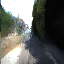}{\caption*{$t=10$}}\endminipage\hfill
    \minipage{0.089\linewidth}\includegraphics[width=\linewidth]{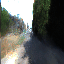}{\caption*{$t=13$}}\endminipage\hfill
    \minipage{0.089\linewidth}\includegraphics[width=\linewidth]{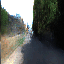}{\caption*{$t=16$}}\endminipage\hfill
    \minipage{0.089\linewidth}\includegraphics[width=\linewidth]{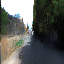}{\caption*{$t=18$}}\endminipage\hfill
    \minipage{0.089\linewidth}\includegraphics[width=\linewidth]{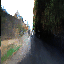}{\caption*{$t=25$}}\endminipage\hfill
  \caption{KITTI Driving Video Prediction. Both GHVAE and SVG' (M=3, K=5) are given the same five context images. Here, a 6-module GHVAE model exhibits performance advantage over SVG' (M=3, K=5).}\label{kitti}
\end{figure}

%% file: human.tex
\begin{figure}[t]
  \centering
    \minipage{0.089\linewidth}\textbf{Ground\\Truth}\endminipage\hfill 
    \minipage{0.089\linewidth}\includegraphics[width=\linewidth]{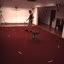}{}\endminipage\hfill
    \minipage{0.089\linewidth}\includegraphics[width=\linewidth]{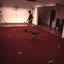}{}\endminipage\hfill
    \minipage{0.089\linewidth}\includegraphics[width=\linewidth]{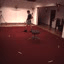}{}\endminipage\hfill
    \minipage{0.089\linewidth}\includegraphics[width=\linewidth]{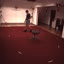}{}\endminipage\hfill
    \minipage{0.089\linewidth}\includegraphics[width=\linewidth]{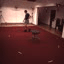}{}\endminipage\hfill
    \minipage{0.089\linewidth}\includegraphics[width=\linewidth]{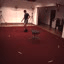}{}\endminipage\hfill
    \minipage{0.089\linewidth}\includegraphics[width=\linewidth]{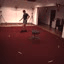}{}\endminipage\hfill
    \minipage{0.089\linewidth}\includegraphics[width=\linewidth]{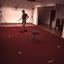}{}\endminipage\hfill
    \minipage{0.089\linewidth}\includegraphics[width=\linewidth]{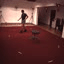}{}\endminipage\hfill
    \minipage{0.089\linewidth}\includegraphics[width=\linewidth]{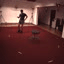}{}\endminipage\hfill
    \minipage{0.089\linewidth}\textbf{GHVAE}\endminipage\hfill 
    \minipage{0.089\linewidth}\includegraphics[width=\linewidth]{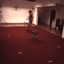}{}\endminipage\hfill
    \minipage{0.089\linewidth}\includegraphics[width=\linewidth]{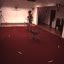}{}\endminipage\hfill
    \minipage{0.089\linewidth}\includegraphics[width=\linewidth]{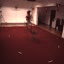}{}\endminipage\hfill
    \minipage{0.089\linewidth}\includegraphics[width=\linewidth]{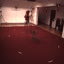}{}\endminipage\hfill
    \minipage{0.089\linewidth}\includegraphics[width=\linewidth]{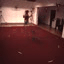}{}\endminipage\hfill
    \minipage{0.089\linewidth}\includegraphics[width=\linewidth]{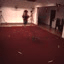}{}\endminipage\hfill
    \minipage{0.089\linewidth}\includegraphics[width=\linewidth]{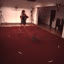}{}\endminipage\hfill
    \minipage{0.089\linewidth}\includegraphics[width=\linewidth]{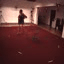}{}\endminipage\hfill
    \minipage{0.089\linewidth}\includegraphics[width=\linewidth]{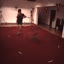}{}\endminipage\hfill
    \minipage{0.089\linewidth}\includegraphics[width=\linewidth]{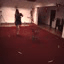}{}\endminipage\hfill
    \minipage{0.089\linewidth}\textbf{SVG' (M=3, K=5)}\endminipage\hfill 
    \minipage{0.089\linewidth}\includegraphics[width=\linewidth]{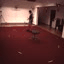}{\caption*{$t=1$}}\endminipage\hfill
    \minipage{0.089\linewidth}\includegraphics[width=\linewidth]{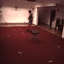}{\caption*{$t=2$}}\endminipage\hfill
    \minipage{0.089\linewidth}\includegraphics[width=\linewidth]{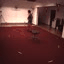}{\caption*{$t=4$}}\endminipage\hfill
    \minipage{0.089\linewidth}\includegraphics[width=\linewidth]{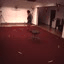}{\caption*{$t=6$}}\endminipage\hfill
    \minipage{0.089\linewidth}\includegraphics[width=\linewidth]{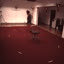}{\caption*{$t=8$}}\endminipage\hfill
    \minipage{0.089\linewidth}\includegraphics[width=\linewidth]{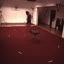}{\caption*{$t=10$}}\endminipage\hfill
    \minipage{0.089\linewidth}\includegraphics[width=\linewidth]{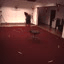}{\caption*{$t=13$}}\endminipage\hfill
    \minipage{0.089\linewidth}\includegraphics[width=\linewidth]{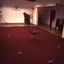}{\caption*{$t=16$}}\endminipage\hfill
    \minipage{0.089\linewidth}\includegraphics[width=\linewidth]{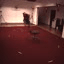}{\caption*{$t=18$}}\endminipage\hfill
    \minipage{0.089\linewidth}\includegraphics[width=\linewidth]{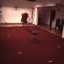}{\caption*{$t=25$}}\endminipage\hfill
  \caption{Human3.6M Video Prediction. Both GHVAE and SVG' (M=3, K=5) are given the same five context images. Here, a 6-module GHVAE model exhibits performance advantage over SVG' (M=3, K=5).}\label{human}
\end{figure}

%% file: cityscapes.tex
\begin{figure}[t]
  \centering
    \minipage{0.089\linewidth}\textbf{Ground\\Truth}\endminipage\hfill 
    \minipage{0.089\linewidth}\includegraphics[width=\linewidth]{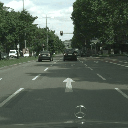}{}\endminipage\hfill
    \minipage{0.089\linewidth}\includegraphics[width=\linewidth]{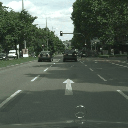}{}\endminipage\hfill
    \minipage{0.089\linewidth}\includegraphics[width=\linewidth]{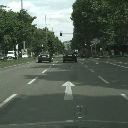}{}\endminipage\hfill
    \minipage{0.089\linewidth}\includegraphics[width=\linewidth]{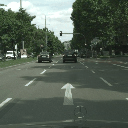}{}\endminipage\hfill
    \minipage{0.089\linewidth}\includegraphics[width=\linewidth]{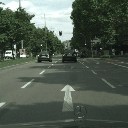}{}\endminipage\hfill
    \minipage{0.089\linewidth}\includegraphics[width=\linewidth]{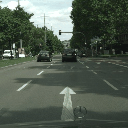}{}\endminipage\hfill
    \minipage{0.089\linewidth}\includegraphics[width=\linewidth]{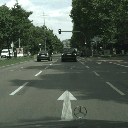}{}\endminipage\hfill
    \minipage{0.089\linewidth}\includegraphics[width=\linewidth]{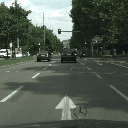}{}\endminipage\hfill
    \minipage{0.089\linewidth}\includegraphics[width=\linewidth]{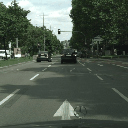}{}\endminipage\hfill
    \minipage{0.089\linewidth}\includegraphics[width=\linewidth]{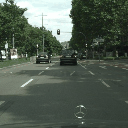}{}\endminipage\hfill
    \minipage{0.089\linewidth}\textbf{GHVAE}\endminipage\hfill 
    \minipage{0.089\linewidth}\includegraphics[width=\linewidth]{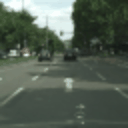}{\caption*{$t=1$}}\endminipage\hfill
    \minipage{0.089\linewidth}\includegraphics[width=\linewidth]{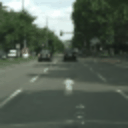}{\caption*{$t=2$}}\endminipage\hfill
    \minipage{0.089\linewidth}\includegraphics[width=\linewidth]{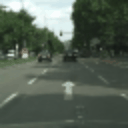}{\caption*{$t=4$}}\endminipage\hfill
    \minipage{0.089\linewidth}\includegraphics[width=\linewidth]{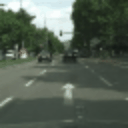}{\caption*{$t=6$}}\endminipage\hfill
    \minipage{0.089\linewidth}\includegraphics[width=\linewidth]{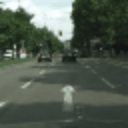}{\caption*{$t=8$}}\endminipage\hfill
    \minipage{0.089\linewidth}\includegraphics[width=\linewidth]{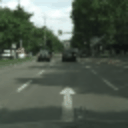}{\caption*{$t=10$}}\endminipage\hfill
    \minipage{0.089\linewidth}\includegraphics[width=\linewidth]{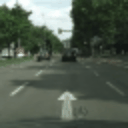}{\caption*{$t=13$}}\endminipage\hfill
    \minipage{0.089\linewidth}\includegraphics[width=\linewidth]{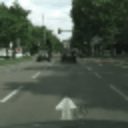}{\caption*{$t=16$}}\endminipage\hfill
    \minipage{0.089\linewidth}\includegraphics[width=\linewidth]{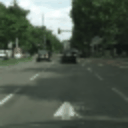}{\caption*{$t=18$}}\endminipage\hfill
    \minipage{0.089\linewidth}\includegraphics[width=\linewidth]{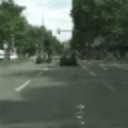}{\caption*{$t=28$}}\endminipage\hfill
  \caption{Cityscapes Driving Video Prediction. Both GHVAE and Hier-VRNN are given the same two context images. Here, a 6-module GHVAE model exhibits performance advantage over Hier-VRNN. Note that this paper directly compares to Hier-VRNN results reported in Castrejon et al.~\cite{Castrejon_2019_ICCV} and does not re-implement the Hier-VRNN algorithm.}\label{cityscapes}
\end{figure}

%% file: sweeping.tex
\begin{figure}[t]
\minipage{0.089\linewidth}\textbf{GHVAE}\endminipage\hfill 
    \minipage{0.089\linewidth}\includegraphics[width=\linewidth]{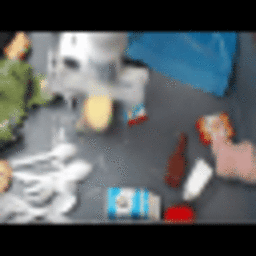}{}\endminipage\hfill
    \minipage{0.089\linewidth}\includegraphics[width=\linewidth]{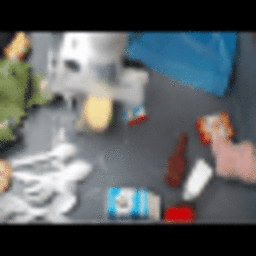}{}\endminipage\hfill
    \minipage{0.089\linewidth}\includegraphics[width=\linewidth]{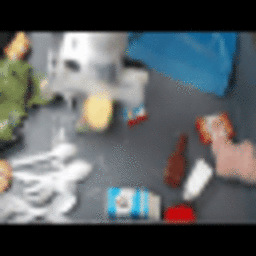}{}\endminipage\hfill
    \minipage{0.089\linewidth}\includegraphics[width=\linewidth]{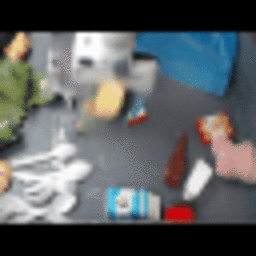}{}\endminipage\hfill
    \minipage{0.089\linewidth}\includegraphics[width=\linewidth]{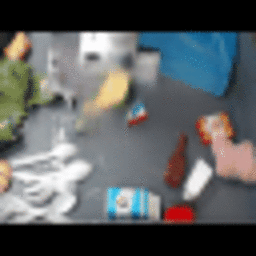}{}\endminipage\hfill
    \minipage{0.089\linewidth}\includegraphics[width=\linewidth]{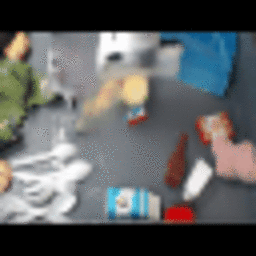}{}\endminipage\hfill
    \minipage{0.089\linewidth}\includegraphics[width=\linewidth]{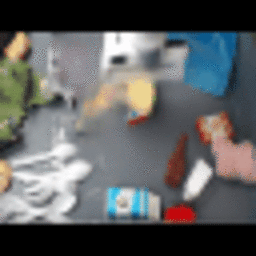}{}\endminipage\hfill
    \minipage{0.089\linewidth}\includegraphics[width=\linewidth]{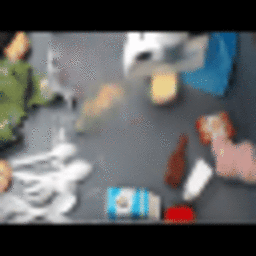}{}\endminipage\hfill
    \minipage{0.089\linewidth}\includegraphics[width=\linewidth]{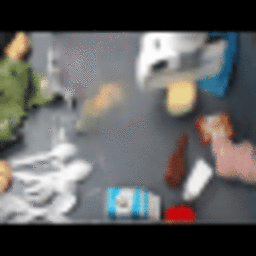}{}\endminipage\hfill
    \minipage{0.089\linewidth}\includegraphics[width=\linewidth]{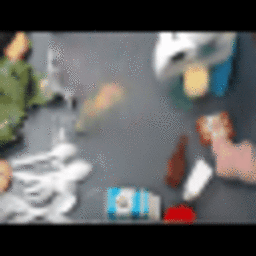}{}\endminipage\hfill\minipage{0.089\linewidth}\textbf{SVG'}\endminipage\hfill 
    \minipage{0.089\linewidth}\includegraphics[width=\linewidth]{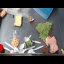}{\caption*{$t=1$}}\endminipage\hfill 
    \minipage{0.089\linewidth}\includegraphics[width=\linewidth]{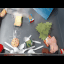}{\caption*{$t=2$}}\endminipage\hfill 
    \minipage{0.089\linewidth}\includegraphics[width=\linewidth]{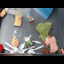}{\caption*{$t=3$}}\endminipage\hfill 
    \minipage{0.089\linewidth}\includegraphics[width=\linewidth]{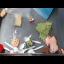}{\caption*{$t=4$}}\endminipage\hfill 
    \minipage{0.089\linewidth}\includegraphics[width=\linewidth]{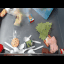}{\caption*{$t=5$}}\endminipage\hfill 
    \minipage{0.089\linewidth}\includegraphics[width=\linewidth]{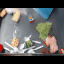}{\caption*{$t=6$}}\endminipage\hfill 
    \minipage{0.089\linewidth}\includegraphics[width=\linewidth]{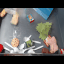}{\caption*{$t=7$}}\endminipage\hfill 
    \minipage{0.089\linewidth}\includegraphics[width=\linewidth]{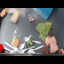}{\caption*{$t=8$}}\endminipage\hfill 
    \minipage{0.089\linewidth}\includegraphics[width=\linewidth]{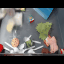}{\caption*{$t=9$}}\endminipage\hfill 
    \minipage{0.089\linewidth}\includegraphics[width=\linewidth]{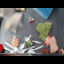}{\caption*{$t=10$}}\endminipage\hfill\\
  \centering
  \caption{Video Prediction in Real-Robot Pick\&Sweep Tasks. Both GHVAE and SVG' are given the same two context images. Here, GHVAE exhibits performance advantage over SVG'. Note that due to our random shooting planning strategy, the rollout length of each method is variable and different in every trial. Kindly see Appendix~\ref{sec:randomshooting}  for more details.}\label{sweeping}
\end{figure}

%% file: wiping.tex
\begin{figure}[t]
  \centering
  \minipage{0.089\linewidth}\textbf{GHVAE}\endminipage
  \minipage{0.089\linewidth}\includegraphics[width=\linewidth]{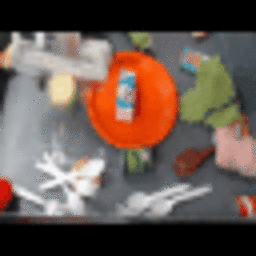}{}\endminipage\hfill
    \minipage{0.089\linewidth}\includegraphics[width=\linewidth]{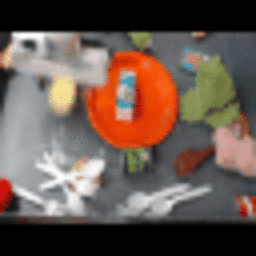}{}\endminipage\hfill
    \minipage{0.089\linewidth}\includegraphics[width=\linewidth]{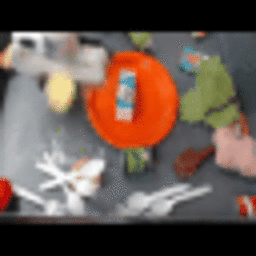}{}\endminipage\hfill
    \minipage{0.089\linewidth}\includegraphics[width=\linewidth]{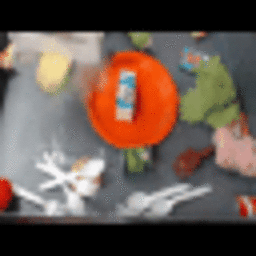}{}\endminipage\hfill
    \minipage{0.089\linewidth}\includegraphics[width=\linewidth]{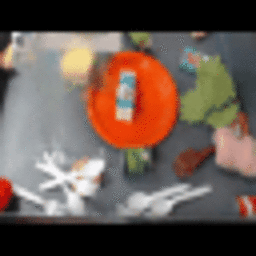}{}\endminipage\hfill
    \minipage{0.089\linewidth}\includegraphics[width=\linewidth]{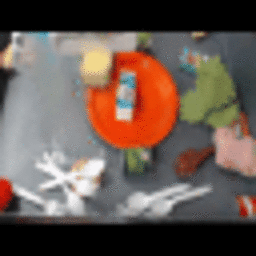}{}\endminipage\hfill
    \minipage{0.089\linewidth}\includegraphics[width=\linewidth]{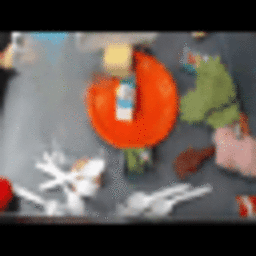}{}\endminipage\hfill
    \minipage{0.089\linewidth}\includegraphics[width=\linewidth]{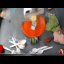}{}\endminipage\hfill
    \minipage{0.089\linewidth}\includegraphics[width=\linewidth]{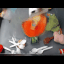}{}\endminipage\hfill
    \minipage{0.089\linewidth}\includegraphics[width=\linewidth]{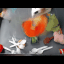}{}\endminipage\hfill\\
    \minipage{0.089\linewidth}\textbf{SVG'}\endminipage\hfill
    \minipage{0.089\linewidth}\includegraphics[width=\linewidth]{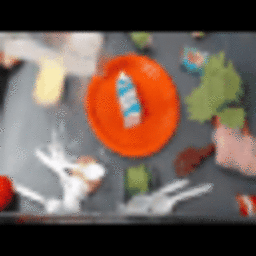}{\caption*{$t=1$}}\endminipage\hfill
    \minipage{0.089\linewidth}\includegraphics[width=\linewidth]{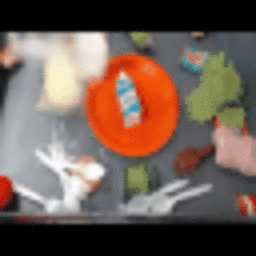}{\caption*{$t=2$}}\endminipage\hfill
    \minipage{0.089\linewidth}\includegraphics[width=\linewidth]{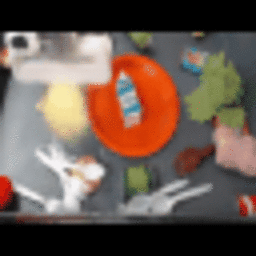}{\caption*{$t=3$}}\endminipage\hfill
    \minipage{0.089\linewidth}\includegraphics[width=\linewidth]{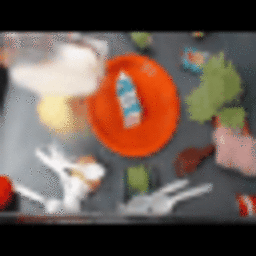}{\caption*{$t=4$}}\endminipage\hfill
    \minipage{0.089\linewidth}\includegraphics[width=\linewidth]{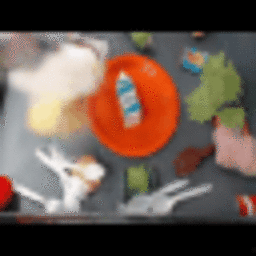}{\caption*{$t=5$}}\endminipage\hfill
    \minipage{0.089\linewidth}\includegraphics[width=\linewidth]{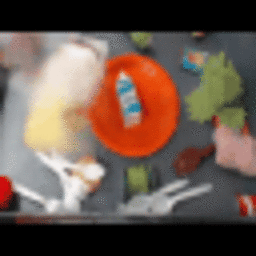}{\caption*{$t=6$}}\endminipage\hfill
    \minipage{0.089\linewidth}\includegraphics[width=\linewidth]{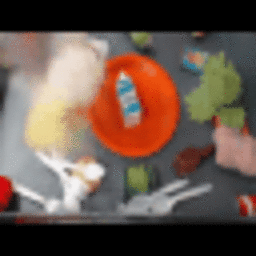}{\caption*{$t=7$}}\endminipage\hfill
    \minipage{0.089\linewidth}\includegraphics[width=\linewidth]{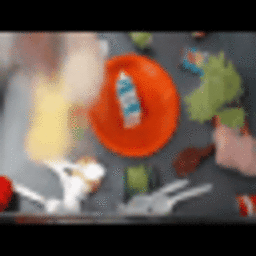}{\caption*{$t=8$}}\endminipage\hfill
    \minipage{0.089\linewidth}\includegraphics[width=\linewidth]{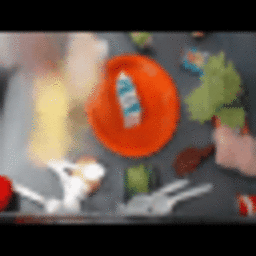}{\caption*{$t=9$}}\endminipage\hfill
    \minipage{0.089\linewidth}\includegraphics[width=\linewidth]{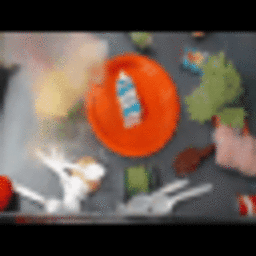}{\caption*{$t=10$}}\endminipage\hfill \\
  \caption{Video Prediction in Real-Robot Pick\&Wipe Tasks. Both GHVAE and SVG' are given the same two context images. Here, GHVAE exhibits performance advantage over SVG'. Note that due to our random shooting planning strategy, the rollout length of each method is variable and different in every trial. Kindly see Appendix~\ref{sec:randomshooting}  for more details.}\label{wiping}
\end{figure}

%% file: sweeping-robot.tex
\begin{figure}[t]
\minipage{0.089\linewidth}\textbf{GHVAE Goal\\ Image}\endminipage\hfill 
    \minipage{0.089\linewidth}\includegraphics[width=\linewidth]{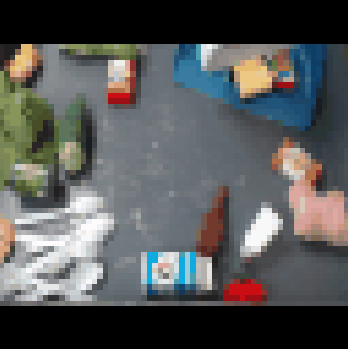}{}\endminipage\hfill
    \minipage{0.089\linewidth}\includegraphics[width=\linewidth]{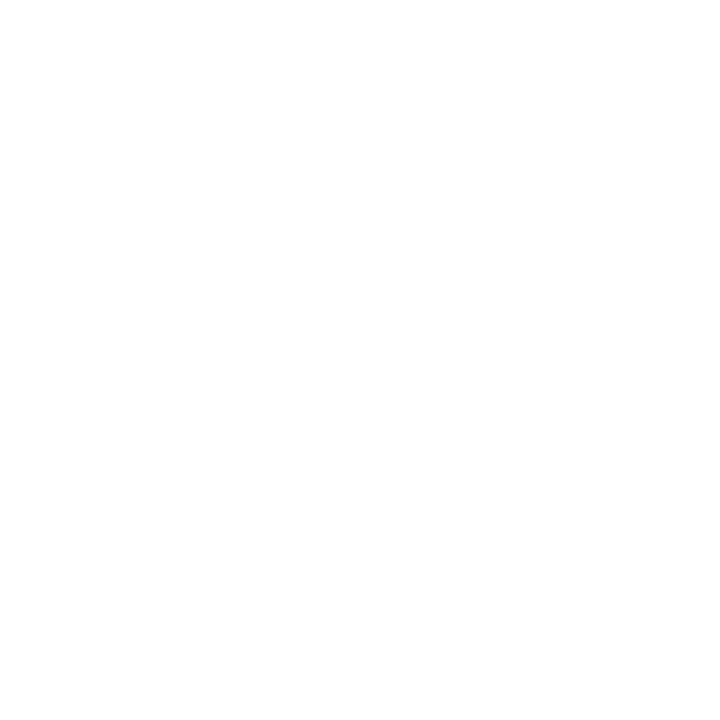}{}\endminipage\hfill
    \minipage{0.089\linewidth}\includegraphics[width=\linewidth]{images/blank.png}{}\endminipage\hfill
    \minipage{0.089\linewidth}\includegraphics[width=\linewidth]{images/blank.png}{}\endminipage\hfill
    \minipage{0.089\linewidth}\includegraphics[width=\linewidth]{images/blank.png}{}\endminipage\hfill
    \minipage{0.089\linewidth}\includegraphics[width=\linewidth]{images/blank.png}{}\endminipage\hfill
    \minipage{0.089\linewidth}\includegraphics[width=\linewidth]{images/blank.png}{}\endminipage\hfill
    \minipage{0.089\linewidth}\includegraphics[width=\linewidth]{images/blank.png}{}\endminipage\hfill
    \minipage{0.089\linewidth}\includegraphics[width=\linewidth]{images/blank.png}{}\endminipage\hfill
    \minipage{0.089\linewidth}\includegraphics[width=\linewidth]{images/blank.png}{}\endminipage\hfill
    
\minipage{0.089\linewidth}\textbf{GHVAE (Success)}\endminipage\hfill 
    \minipage{0.089\linewidth}\includegraphics[width=\linewidth]{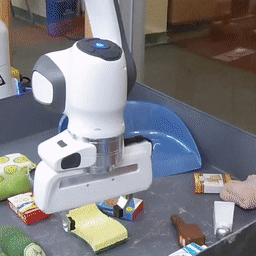}{}\endminipage\hfill
    \minipage{0.089\linewidth}\includegraphics[width=\linewidth]{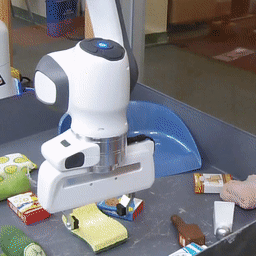}{}\endminipage\hfill
    \minipage{0.089\linewidth}\includegraphics[width=\linewidth]{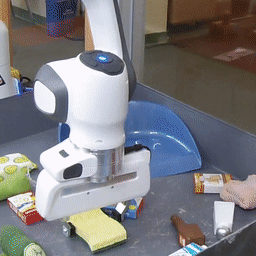}{}\endminipage\hfill
    \minipage{0.089\linewidth}\includegraphics[width=\linewidth]{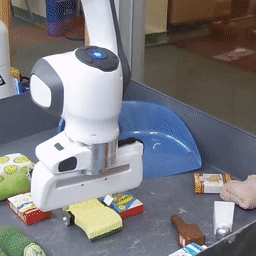}{}\endminipage\hfill
    \minipage{0.089\linewidth}\includegraphics[width=\linewidth]{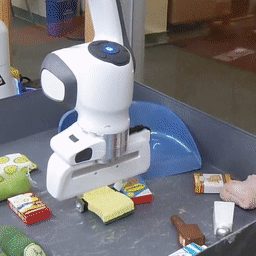}{}\endminipage\hfill
    \minipage{0.089\linewidth}\includegraphics[width=\linewidth]{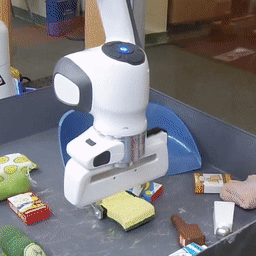}{}\endminipage\hfill
    \minipage{0.089\linewidth}\includegraphics[width=\linewidth]{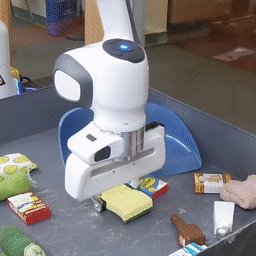}{}\endminipage\hfill
    \minipage{0.089\linewidth}\includegraphics[width=\linewidth]{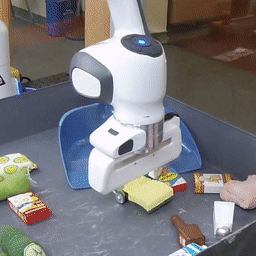}{}\endminipage\hfill
    \minipage{0.089\linewidth}\includegraphics[width=\linewidth]{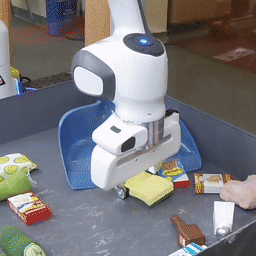}{}\endminipage\hfill
    \minipage{0.089\linewidth}\includegraphics[width=\linewidth]{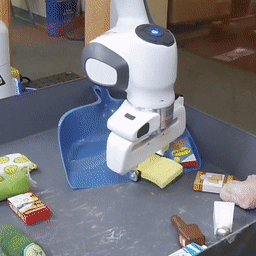}{}\endminipage\hfill
\minipage{0.089\linewidth}\textbf{SVG' Goal\\ Image}\endminipage\hfill 
    \minipage{0.089\linewidth}\includegraphics[width=\linewidth]{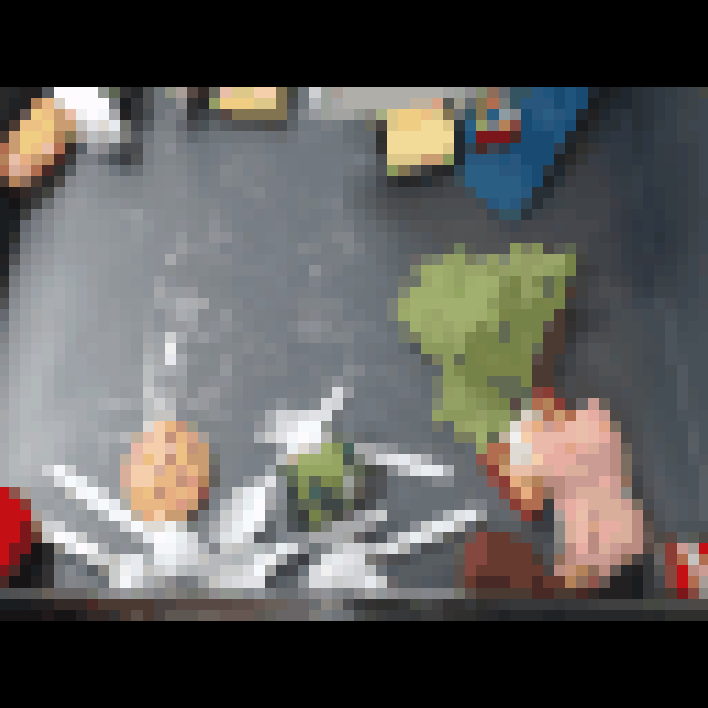}{}\endminipage\hfill
    \minipage{0.089\linewidth}\includegraphics[width=\linewidth]{images/blank.png}{}\endminipage\hfill
    \minipage{0.089\linewidth}\includegraphics[width=\linewidth]{images/blank.png}{}\endminipage\hfill
    \minipage{0.089\linewidth}\includegraphics[width=\linewidth]{images/blank.png}{}\endminipage\hfill
    \minipage{0.089\linewidth}\includegraphics[width=\linewidth]{images/blank.png}{}\endminipage\hfill
    \minipage{0.089\linewidth}\includegraphics[width=\linewidth]{images/blank.png}{}\endminipage\hfill
    \minipage{0.089\linewidth}\includegraphics[width=\linewidth]{images/blank.png}{}\endminipage\hfill
    \minipage{0.089\linewidth}\includegraphics[width=\linewidth]{images/blank.png}{}\endminipage\hfill
    \minipage{0.089\linewidth}\includegraphics[width=\linewidth]{images/blank.png}{}\endminipage\hfill
    \minipage{0.089\linewidth}\includegraphics[width=\linewidth]{images/blank.png}{}\endminipage\hfill
    \minipage{0.089\linewidth}\textbf{SVG' (Failed)}\endminipage\hfill 
    \minipage{0.089\linewidth}\includegraphics[width=\linewidth]{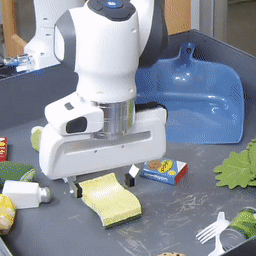}\endminipage\hfill 
    \minipage{0.089\linewidth}\includegraphics[width=\linewidth]{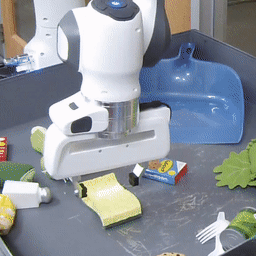}\endminipage\hfill
    \minipage{0.089\linewidth}\includegraphics[width=\linewidth]{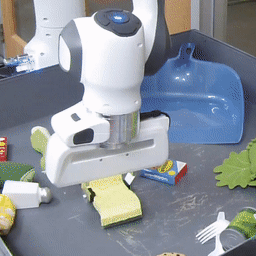}\endminipage\hfill
    \minipage{0.089\linewidth}\includegraphics[width=\linewidth]{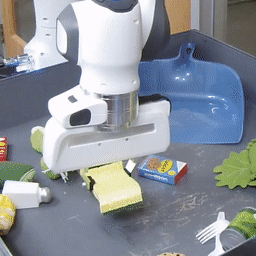}\endminipage\hfill
    \minipage{0.089\linewidth}\includegraphics[width=\linewidth]{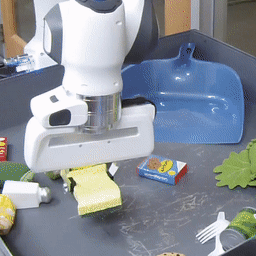}\endminipage\hfill
    \minipage{0.089\linewidth}\includegraphics[width=\linewidth]{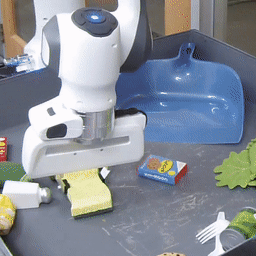}\endminipage\hfill
    \minipage{0.089\linewidth}\includegraphics[width=\linewidth]{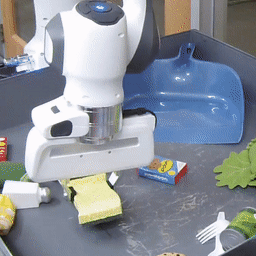}\endminipage\hfill
    \minipage{0.089\linewidth}\includegraphics[width=\linewidth]{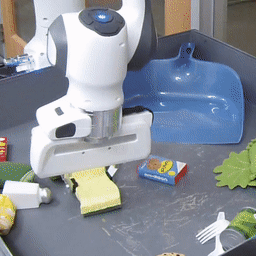}\endminipage\hfill
    \minipage{0.089\linewidth}\includegraphics[width=\linewidth]{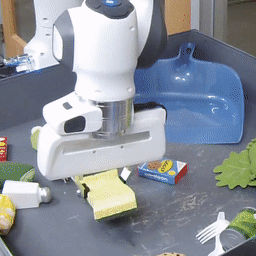}\endminipage\hfill
    \minipage{0.089\linewidth}\includegraphics[width=\linewidth]{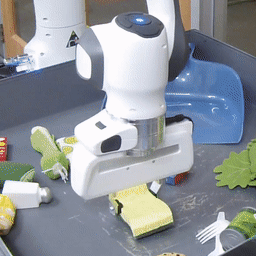}\endminipage\hfill\\
  \centering
  \caption{Real-Robot Task Execution in Pick\&Sweep Experiments. Here, a 6-module GHVAE model exhibits more frequent successes than SVG'.}\label{sweeping-robot}
\end{figure}

%% file: wiping-robot.tex
\begin{figure}[t]
\minipage{0.089\linewidth}\textbf{GHVAE Goal\\ Image}\endminipage\hfill 
    \minipage{0.089\linewidth}\includegraphics[width=\linewidth]{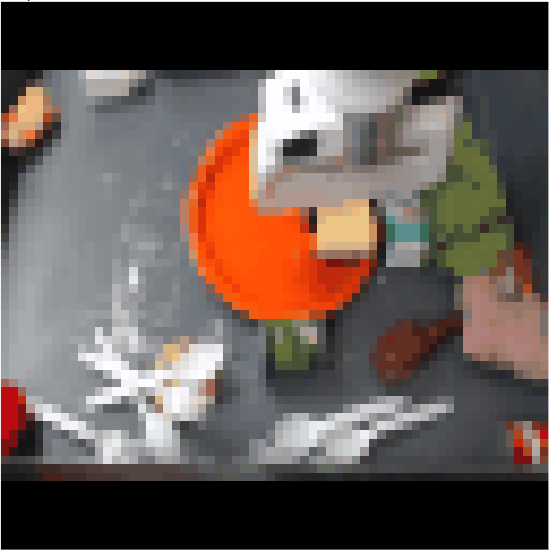}{}\endminipage\hfill
    \minipage{0.089\linewidth}\includegraphics[width=\linewidth]{images/blank.png}{}\endminipage\hfill
    \minipage{0.089\linewidth}\includegraphics[width=\linewidth]{images/blank.png}{}\endminipage\hfill
    \minipage{0.089\linewidth}\includegraphics[width=\linewidth]{images/blank.png}{}\endminipage\hfill
    \minipage{0.089\linewidth}\includegraphics[width=\linewidth]{images/blank.png}{}\endminipage\hfill
    \minipage{0.089\linewidth}\includegraphics[width=\linewidth]{images/blank.png}{}\endminipage\hfill
    \minipage{0.089\linewidth}\includegraphics[width=\linewidth]{images/blank.png}{}\endminipage\hfill
    \minipage{0.089\linewidth}\includegraphics[width=\linewidth]{images/blank.png}{}\endminipage\hfill
    \minipage{0.089\linewidth}\includegraphics[width=\linewidth]{images/blank.png}{}\endminipage\hfill
    \minipage{0.089\linewidth}\includegraphics[width=\linewidth]{images/blank.png}{}\endminipage\hfill
    
\minipage{0.089\linewidth}\textbf{GHVAE (Success)}\endminipage\hfill 
    \minipage{0.089\linewidth}\includegraphics[width=\linewidth]{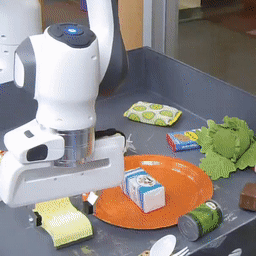}{}\endminipage\hfill
    \minipage{0.089\linewidth}\includegraphics[width=\linewidth]{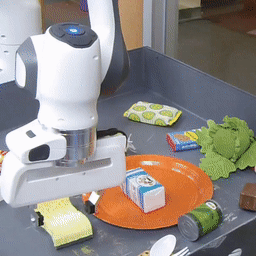}{}\endminipage\hfill
    \minipage{0.089\linewidth}\includegraphics[width=\linewidth]{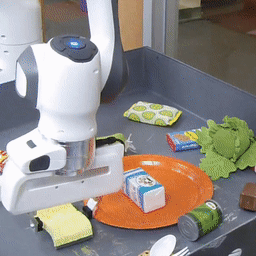}{}\endminipage\hfill
    \minipage{0.089\linewidth}\includegraphics[width=\linewidth]{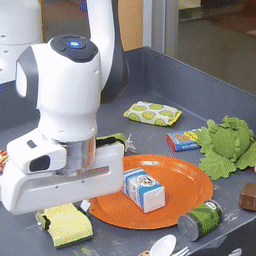}{}\endminipage\hfill
    \minipage{0.089\linewidth}\includegraphics[width=\linewidth]{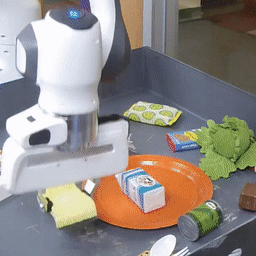}{}\endminipage\hfill
    \minipage{0.089\linewidth}\includegraphics[width=\linewidth]{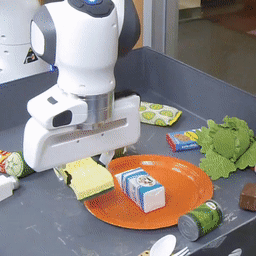}{}\endminipage\hfill
    \minipage{0.089\linewidth}\includegraphics[width=\linewidth]{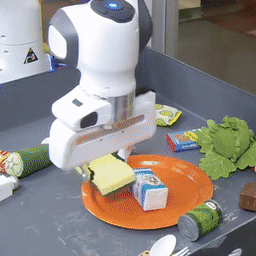}{}\endminipage\hfill
    \minipage{0.089\linewidth}\includegraphics[width=\linewidth]{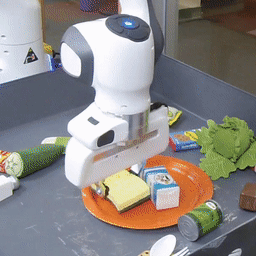}{}\endminipage\hfill
    \minipage{0.089\linewidth}\includegraphics[width=\linewidth]{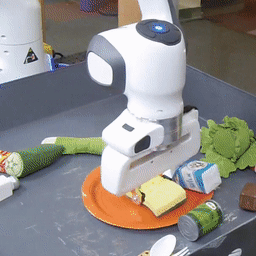}{}\endminipage\hfill
    \minipage{0.089\linewidth}\includegraphics[width=\linewidth]{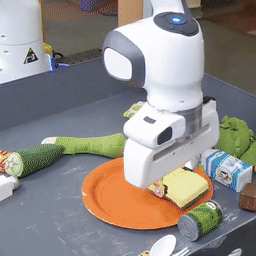}{}\endminipage\hfill
\minipage{0.089\linewidth}\textbf{SVG' Goal\\ Image}\endminipage\hfill 
    \minipage{0.089\linewidth}\includegraphics[width=\linewidth]{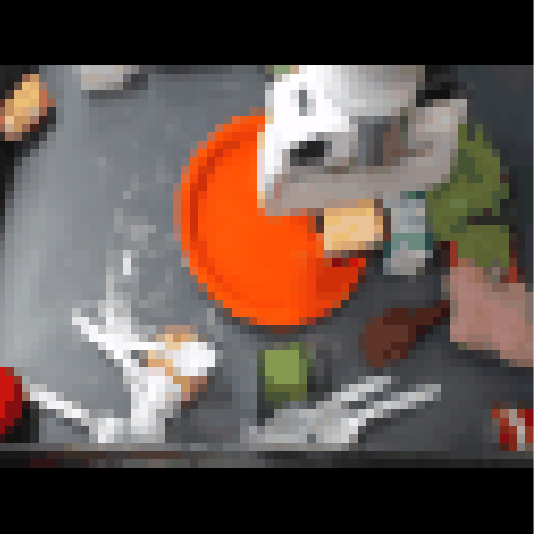}{}\endminipage\hfill
    \minipage{0.089\linewidth}\includegraphics[width=\linewidth]{images/blank.png}{}\endminipage\hfill
    \minipage{0.089\linewidth}\includegraphics[width=\linewidth]{images/blank.png}{}\endminipage\hfill
    \minipage{0.089\linewidth}\includegraphics[width=\linewidth]{images/blank.png}{}\endminipage\hfill
    \minipage{0.089\linewidth}\includegraphics[width=\linewidth]{images/blank.png}{}\endminipage\hfill
    \minipage{0.089\linewidth}\includegraphics[width=\linewidth]{images/blank.png}{}\endminipage\hfill
    \minipage{0.089\linewidth}\includegraphics[width=\linewidth]{images/blank.png}{}\endminipage\hfill
    \minipage{0.089\linewidth}\includegraphics[width=\linewidth]{images/blank.png}{}\endminipage\hfill
    \minipage{0.089\linewidth}\includegraphics[width=\linewidth]{images/blank.png}{}\endminipage\hfill
    \minipage{0.089\linewidth}\includegraphics[width=\linewidth]{images/blank.png}{}\endminipage\hfill
    \minipage{0.089\linewidth}\textbf{SVG' (Failed)}\endminipage\hfill 
    \minipage{0.089\linewidth}\includegraphics[width=\linewidth]{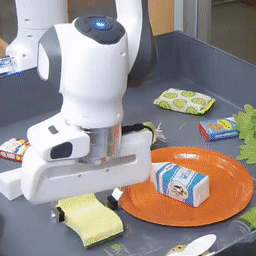}\endminipage\hfill 
    \minipage{0.089\linewidth}\includegraphics[width=\linewidth]{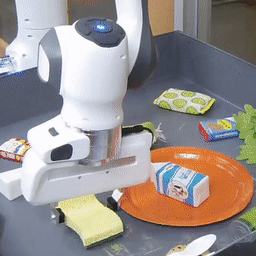}\endminipage\hfill
    \minipage{0.089\linewidth}\includegraphics[width=\linewidth]{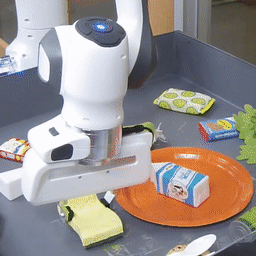}\endminipage\hfill
    \minipage{0.089\linewidth}\includegraphics[width=\linewidth]{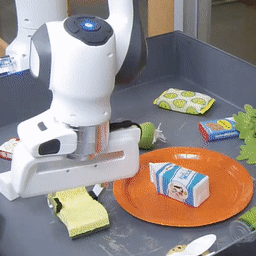}\endminipage\hfill
    \minipage{0.089\linewidth}\includegraphics[width=\linewidth]{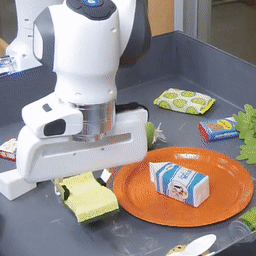}\endminipage\hfill
    \minipage{0.089\linewidth}\includegraphics[width=\linewidth]{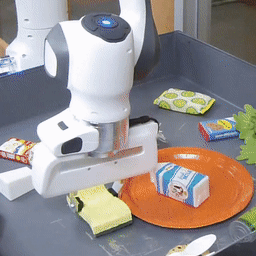}\endminipage\hfill
    \minipage{0.089\linewidth}\includegraphics[width=\linewidth]{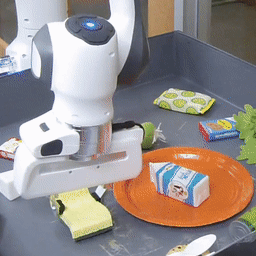}\endminipage\hfill
    \minipage{0.089\linewidth}\includegraphics[width=\linewidth]{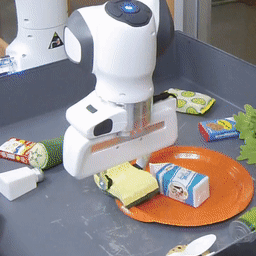}\endminipage\hfill
    \minipage{0.089\linewidth}\includegraphics[width=\linewidth]{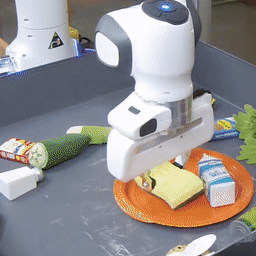}\endminipage\hfill
    \minipage{0.089\linewidth}\includegraphics[width=\linewidth]{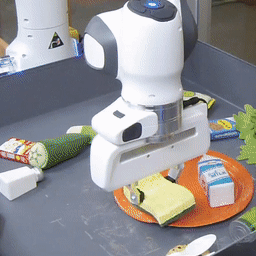}\endminipage\hfill\\
  \centering
  \caption{Real-Robot Task Execution in Pick\&Wipe Experiments. Here, a 6-module GHVAE model exhibits more frequent successes than SVG'.}\label{wiping-robot}
\end{figure}

%% file: task-env.tex
\begin{figure}[t]
  \centering
    \minipage{0.165\linewidth}\textbf{Pick\&Sweep Tasks}\endminipage\hfill 
    \minipage{0.165\linewidth}\includegraphics[width=\linewidth]{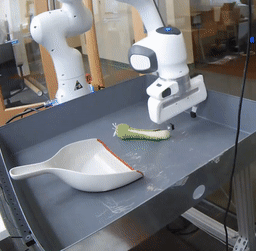}{}\endminipage\hfill
    \minipage{0.165\linewidth}\includegraphics[width=\linewidth]{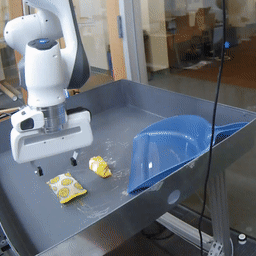}{}\endminipage\hfill
    \minipage{0.165\linewidth}\includegraphics[width=\linewidth]{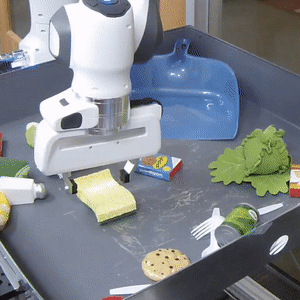}{}\endminipage\hfill
    \minipage{0.165\linewidth}\includegraphics[width=\linewidth]{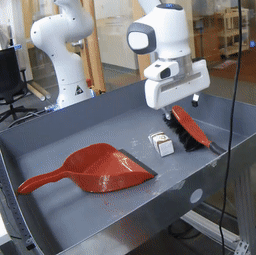}{}\endminipage\hfill
    \minipage{0.165\linewidth}\includegraphics[width=\linewidth]{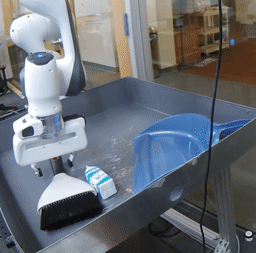}{}\endminipage\hfill
    \minipage{0.165\linewidth}\textbf{Pick\&Wipe Tasks}\endminipage\hfill 
    \minipage{0.165\linewidth}\includegraphics[width=\linewidth]{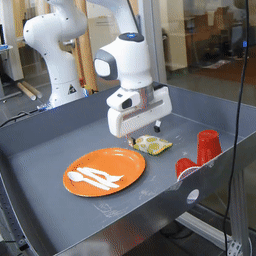}{}\endminipage\hfill
    \minipage{0.165\linewidth}\includegraphics[width=\linewidth]{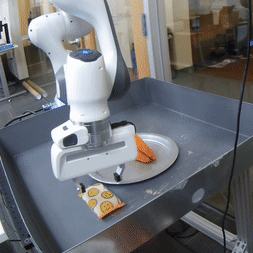}{}\endminipage\hfill
    \minipage{0.165\linewidth}\includegraphics[width=\linewidth]{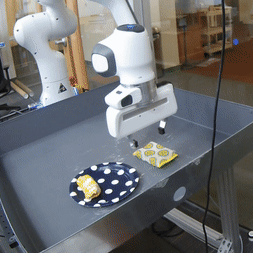}{}\endminipage\hfill
    \minipage{0.165\linewidth}\includegraphics[width=\linewidth]{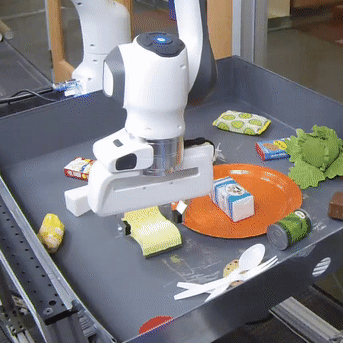}{}\endminipage\hfill
    \minipage{0.165\linewidth}\includegraphics[width=\linewidth]{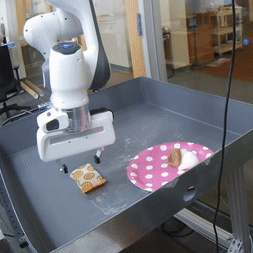}{}\endminipage\hfill
  \caption{Sample Real-Robot Evaluation Tasks}\label{fig:env}
\end{figure}
\begin{figure}[t]
  \centering
    \minipage{0.165\linewidth}\textbf{Sample\\ Training\\ Environment}\endminipage\hfill 
    \minipage{0.165\linewidth}\includegraphics[width=\linewidth]{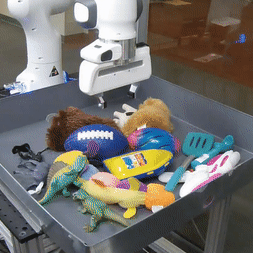}{}\endminipage\hfill
    \minipage{0.165\linewidth}\includegraphics[width=\linewidth]{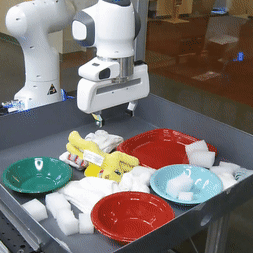}{}\endminipage\hfill
    \minipage{0.165\linewidth}\includegraphics[width=\linewidth]{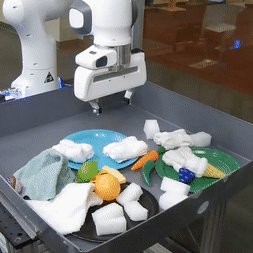}{}\endminipage\hfill
    \minipage{0.165\linewidth}\includegraphics[width=\linewidth]{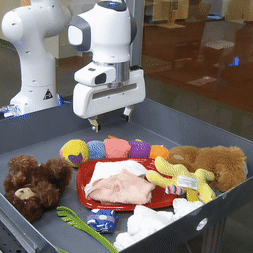}{}\endminipage\hfill
    \minipage{0.165\linewidth}\includegraphics[width=\linewidth]{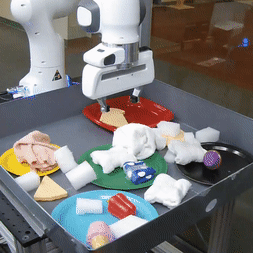}{}\endminipage\hfill
  \caption{\textbf{Representative Real-Robot Training Environment.} Note that all objects used during training are excluded from evaluation. The 5000-video training data for both the Pick\&Sweep and the Pick\&Wipe tasks are the same.}\label{fig:train-env}
\end{figure}

%% file: failure.tex
\begin{figure}[t]
  \centering
    \minipage{0.089\linewidth}\textbf{Ground\\Truth}\endminipage\hfill 
    \minipage{0.089\linewidth}\includegraphics[width=\linewidth]{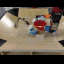}{}\endminipage\hfill
    \minipage{0.089\linewidth}\includegraphics[width=\linewidth]{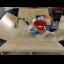}{}\endminipage\hfill
    \minipage{0.089\linewidth}\includegraphics[width=\linewidth]{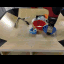}{}\endminipage\hfill
    \minipage{0.089\linewidth}\includegraphics[width=\linewidth]{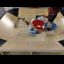}{}\endminipage\hfill
    \minipage{0.089\linewidth}\includegraphics[width=\linewidth]{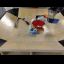}{}\endminipage\hfill
    \minipage{0.089\linewidth}\includegraphics[width=\linewidth]{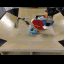}{}\endminipage\hfill
    \minipage{0.089\linewidth}\includegraphics[width=\linewidth]{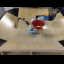}{}\endminipage\hfill
    \minipage{0.089\linewidth}\includegraphics[width=\linewidth]{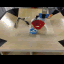}{}\endminipage\hfill
    \minipage{0.089\linewidth}\includegraphics[width=\linewidth]{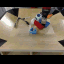}{}\endminipage\hfill
    \minipage{0.089\linewidth}\includegraphics[width=\linewidth]{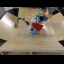}{}\endminipage\hfill
    \minipage{0.089\linewidth}\textbf{GHVAE}\endminipage\hfill 
    \minipage{0.089\linewidth}\includegraphics[width=\linewidth]{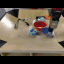}{}\endminipage\hfill
    \minipage{0.089\linewidth}\includegraphics[width=\linewidth]{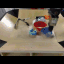}{}\endminipage\hfill
    \minipage{0.089\linewidth}\includegraphics[width=\linewidth]{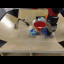}{}\endminipage\hfill
    \minipage{0.089\linewidth}\includegraphics[width=\linewidth]{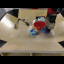}{}\endminipage\hfill
    \minipage{0.089\linewidth}\includegraphics[width=\linewidth]{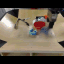}{}\endminipage\hfill
    \minipage{0.089\linewidth}\includegraphics[width=\linewidth]{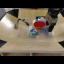}{}\endminipage\hfill
    \minipage{0.089\linewidth}\includegraphics[width=\linewidth]{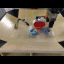}{}\endminipage\hfill
    \minipage{0.089\linewidth}\includegraphics[width=\linewidth]{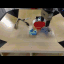}{}\endminipage\hfill
    \minipage{0.089\linewidth}\includegraphics[width=\linewidth]{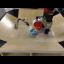}{}\endminipage\hfill
    \minipage{0.089\linewidth}\includegraphics[width=\linewidth]{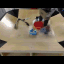}{}\endminipage\hfill
  \caption{Failure case for a 6-module GHVAE model on RoboNet. In this case, the GHVAE model failed to accurately track the movement of the blue bowl. This indicates that the GHVAE model is still slightly underfitting on RoboNet. We hypothesize that training an 8-module, 10-module or 12-module GHVAE model will resolve such failure case.}\label{failure}
\end{figure}